%% file: main.tex
\documentclass[10pt,journal,compsoc]{IEEEtran}
\input{math_symbols.tex}

\usepackage[utf8]{inputenc}
\usepackage[numbers,sort&compress]{natbib}
\usepackage{graphicx}%
\usepackage{multirow}%
\usepackage{amsmath,amssymb,amsfonts}%
\usepackage{booktabs}%
\usepackage{url}
\usepackage{caption}
\usepackage[colorlinks,linkcolor=black,anchorcolor=black,citecolor=black]{hyperref}
\usepackage{makecell}
\usepackage{balance}
\usepackage{ragged2e}
\usepackage{colortbl}
\usepackage{tcolorbox}
\usepackage{todonotes}

\newcommand{\mypara}{\vspace{5pt}\noindent\textbf}

\ifCLASSINFOpdf
\else
\fi

\begin{document}
%
\title{Large Language Models for Information Retrieval: A Survey}
%
%
%
%

\author{
Yutao Zhu, Huaying Yuan, Shuting Wang, Jiongnan Liu, Wenhan Liu, Chenlong Deng \\ Haonan Chen, Zheng Liu, Zhicheng Dou, and Ji-Rong Wen
\thanks{All authors except Zheng Liu are from Gaoling School of Artificial Intelligence and School of Information, Renmin University of China. Zheng Liu is from Beijing Academy of Artificial Intelligence, China. \\
Contact e-mail: yutaozhu94@gmail.com, dou@ruc.edu.cn
}
}

\IEEEtitleabstractindextext{%
\begin{justify}
\begin{abstract}
As a primary means of information acquisition, information retrieval (IR) systems, such as search engines, have integrated themselves into our daily lives. These systems also serve as components of dialogue, question-answering, and recommender systems. The trajectory of IR has evolved dynamically from its origins in term-based methods to its integration with advanced neural models. While the neural models excel at capturing complex contextual signals and semantic nuances, thereby reshaping the IR landscape, they still face challenges such as data scarcity, interpretability, and the generation of contextually plausible yet potentially inaccurate responses. This evolution requires a combination of both traditional methods (such as term-based sparse retrieval methods with rapid response) and modern neural architectures (such as language models with powerful language understanding capacity). Meanwhile, the emergence of large language models (LLMs), typified by ChatGPT and GPT-4, has revolutionized natural language processing due to their remarkable language understanding, generation, generalization, and reasoning abilities. Consequently, recent research has sought to leverage LLMs to improve IR systems. Given the rapid evolution of this research trajectory, it is necessary to consolidate existing methodologies and provide nuanced insights through a comprehensive overview. In this survey, we delve into the confluence of LLMs and IR systems, including crucial aspects such as query rewriters, retrievers, rerankers, and readers. Additionally, we explore promising directions, such as search agents, within this expanding field.
\end{abstract}
\end{justify}

}

\maketitle

\IEEEdisplaynontitleabstractindextext

%
\IEEEpeerreviewmaketitle

\input{S1-introduction}
\input{S2-background}

\input{S3-query-rewriter}
\input{S4-retriever}
\input{S5-reranker}

\input{S6-reader}

\input{S7-agent}

\input{S8-future-direction}
\input{S9-conclusion}


%







\balance
\input{main.bbl}
\end{document}

%% file: math_symbols.tex
\usepackage{amsmath,amsfonts}

\newcommand{\ie}{\textit{i.e.}}
\newcommand{\eg}{\textit{e.g.}}

%% file: S1-introduction.tex
\IEEEraisesectionheading{\section{Introduction}\label{sec:introduction}}

%
%
%
%


\begin{figure*}[t]
    \centering
    \includegraphics[width=.8\linewidth]{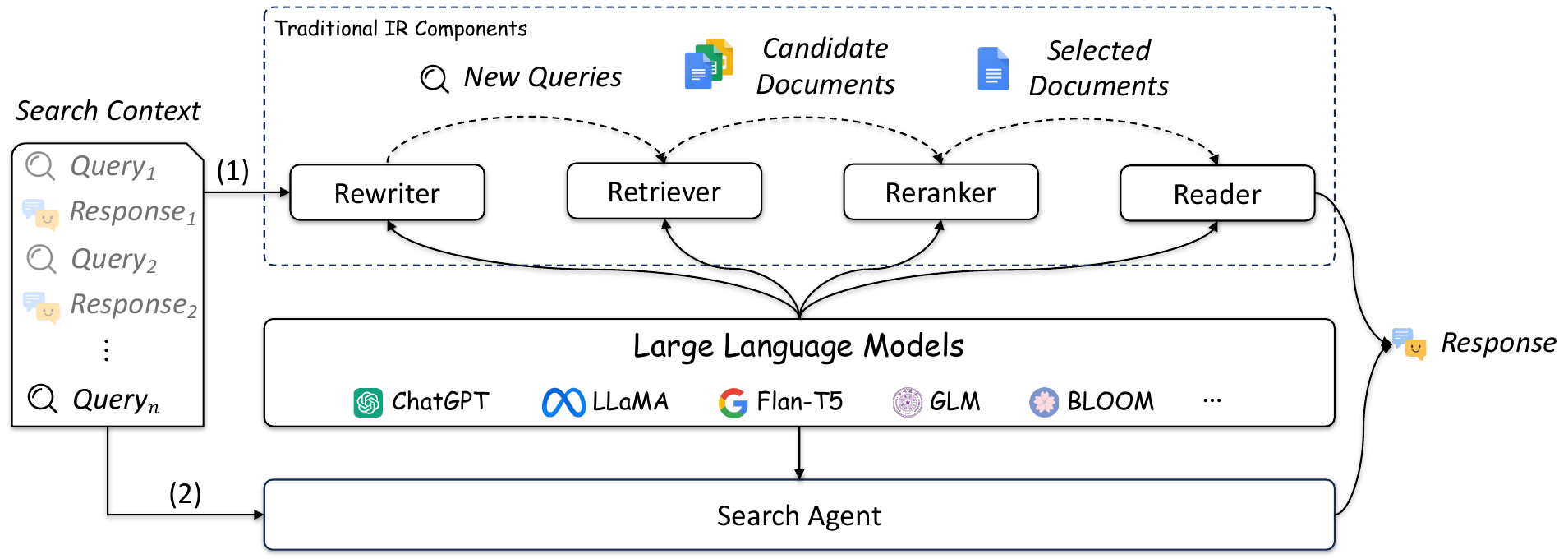}
    \caption{Overview of existing studies that apply LLMs into IR. (1) LLMs can be used to enhance traditional IR components, such as query rewriter, retriever, reranker, and reader. (2) LLMs can also be used as search agents to perform multiple IR tasks.}
    \label{fig:overview}
\end{figure*}

\IEEEPARstart{I}{nformation} access is one of the fundamental daily needs of human beings. To fulfill the need for rapid acquisition of desired information, various information retrieval (IR) systems have been developed~\cite{DBLP:conf/acl/WuWXZL17,xiaoice,dpr,DBLP:journals/csur/DattaJLW08}. Prominent examples include search engines such as Google, Bing, and Baidu, which serve as IR systems on the Internet, adept at retrieving relevant web pages in response to user queries, and provide convenient and efficient access to information on the Internet. It is worth noting that IR extends beyond web page retrieval. In dialogue systems (chatbots)~\cite{DBLP:conf/acl/WuWXZL17,DBLP:conf/emnlp/YuanZLLZHH19,DBLP:conf/ecir/ZhuNZDD21,DBLP:conf/sigir/ZhuNZDJD21,DBLP:journals/corr/abs-2108-07935}, such as Microsoft Xiaoice~\cite{xiaoice}, Apple Siri,\footnote{Apple Siri, \url{https://www.apple.com/siri/}} and Google Assistant,\footnote{Google Assistant, \url{https://assistant.google.com/}} IR systems play a crucial role in retrieving appropriate responses to user input utterances, thereby producing natural and fluent human-machine conversations. Similarly, in question-answering systems~\cite{dpr,rocketqa}, IR systems are employed to select relevant clues essential for addressing user questions effectively. In image search engines~\cite{DBLP:journals/csur/DattaJLW08}, IR systems excel at returning images that align with user input queries. Given the exponential growth of information, research and industry have become increasingly interested in the development of effective IR systems.

The core function of an IR system is retrieval, which aims to determine the relevance between a user-issued query and the content to be retrieved, including various types of information such as texts, images, music, and more. For the scope of this survey, we concentrate solely on reviewing those text retrieval systems, in which query-document relevance is commonly measured by their matching score.\footnote{The term ``document'' will henceforth refer to any text-based content subject to retrieve, including both long articles and short passages.} Given that IR systems operate on extensive repositories, the efficiency of retrieval algorithms becomes of paramount importance. To improve the user experience, the retrieval performance is enhanced from both the upstream (query reformulation) and downstream (reranking and reading) perspectives. As an upstream technique, query reformulation is designed to refine user queries so that they are more effective at retrieving relevant documents~\cite{DBLP:journals/jiis/ArensKS96,DBLP:conf/cikm/HuangE09}. With the recent surge in the popularity of conversational search, this technique has received increasing attention. On the downstream side, reranking approaches are developed to further adjust the document ranking~\cite{monobert,monot5,coca}. In contrast to the retrieval stage, reranking is performed only on a limited set of relevant documents, already retrieved by the retriever. Under this circumstance, the emphasis is placed on achieving higher performance rather than keeping higher efficiency, allowing for the application of more complex approaches in the reranking process. Additionally, reranking can accommodate other specific requirements, such as personalization~\cite{DBLP:conf/sigir/TeevanDH05,DBLP:conf/sigir/BennettWCDBBC12,DBLP:conf/cikm/GeDJNW18,DBLP:conf/cikm/ZhouD0W21} and diversification~\cite{DBLP:conf/sigir/CarbonellG98,DBLP:conf/wsdm/AgrawalGHI09,DBLP:conf/sigir/LiuDWLW20,DBLP:conf/sigir/SuDZQW21}. Following the retrieval and reranking stages, a reading component is incorporated to summarize the retrieved documents and deliver a concise document to users~\cite{RETRO,webgpt}. While traditional IR systems typically require users to gather and organize relevant information themselves; however, the reading component is an integral part of new IR systems such as New Bing,\footnote{New Bing, \url{https://www.bing.com/new}} improving users' browsing experience and saving valuable time.

The trajectory of IR has traversed a dynamic evolution, transitioning from its origins in term-based methods to the integration of neural models. Initially, IR was anchored in term-based methods~\cite{DBLP:books/mg/SaltonG83} and Boolean logic, focusing on keyword matching for document retrieval. The paradigm gradually shifted with the introduction of vector space models~\cite{vector_space}, unlocking the potential to capture nuanced semantic relationships between terms. This progression continued with statistical language models~\cite{DBLP:conf/cikm/SongC99,tfidf}, refining relevance estimation through contextual and probabilistic considerations. The influential BM25 algorithm~\cite{DBLP:conf/trec/RobertsonWJHG94} played an important role during this phase, revolutionizing relevance ranking by accounting for term frequency and document length variations. The most recent chapter in IR's journey is marked by the ascendancy of neural models~\cite{DBLP:conf/cikm/GuoFAC16,dpr,ance,bert4ranking}. These models excel at capturing intricate contextual cues and semantic nuances, reshaping the landscape of IR. However, these neural models still face challenges such as data scarcity, interpretability, and the potential generation of plausible yet inaccurate responses. Thus, the evolution of IR continues to be a journey of balancing traditional strengths (such as the BM25 algorithm's high efficiency) with the remarkable capability (such as semantic understanding) brought about by modern neural architectures.

Large language models (LLMs) have recently emerged as transformative forces across various research fields, such as natural language processing (NLP)~\cite{gpt-2,gpt-3,llama}, recommender systems~\cite{DBLP:journals/corr/abs-2305-07001,DBLP:journals/corr/abs-2305-08845,DBLP:journals/corr/abs-2306-10933,DBLP:journals/corr/abs-2307-02046}, finance~\cite{bloomberggpt}, and even molecule discovery~\cite{DBLP:journals/corr/abs-2306-06615}. These cutting-edge LLMs are primarily based on the Transformer architecture and undergo extensive pre-training on diverse textual sources, including web pages, research articles, books, and codes. As their scale continues to expand (including both model size and data volume), LLMs have demonstrated remarkable advances in their capabilities. On the one hand, LLMs have exhibited unprecedented proficiency in language understanding and generation, resulting in responses that are more human-like and better aligned with human intentions. On the other hand, the larger LLMs have shown impressive emergent abilities when dealing with complex tasks~\cite{emergent_ability}, such as generalization and reasoning skills. 
Leveraging the impressive power of LLMs can undoubtedly improve the performance of IR systems. By incorporating these advanced language models, IR systems can provide users with more accurate responses, ultimately reshaping the landscape of information access and retrieval.

Initial efforts have been made to utilize the potential of LLMs in the development of novel IR systems. Notably, in terms of practical applications, New Bing is designed to improve the users' experience of using search engines by extracting information from disparate web pages and condensing it into concise summaries that serve as responses to user-generated queries. In the research community, LLMs have proven useful within specific modules of IR systems (such as retrievers), thereby enhancing the overall performance of these systems. Due to the rapid evolution of LLM-enhanced IR systems, it is essential to comprehensively review their most recent advancements and challenges. 

Our survey provides an insightful exploration of the intersection between LLMs and IR systems, covering key perspectives such as query rewriters, retrievers, rerankers, and readers (as shown in Figure~\ref{fig:overview}).\footnote{As yet, there has not been a formal definition for LLMs. In this paper, we mainly focus on models with more than 1B parameters. We also notice that some methods do not rely on such strictly defined LLMs, but due to their representativeness, we still include an introduction to them in this survey.} We also include some recent studies that leverage LLMs as search agents to perform various IR tasks. This analysis enhances our understanding of LLMs' potential and limitations in advancing the IR field. For this survey, we create a Github repository by collecting the relevant papers and resources about applying LLMs for IR tasks (LLM4IR).\footnote{\url{https://github.com/RUC-NLPIR/LLM4IR-Survey}} We will continue to update the repository with newer papers. This survey will also be periodically updated according to the development of this area. We notice that there are several surveys for PLMs, LLMs, and their applications (\eg, AIGC or recommender systems)~\cite{DBLP:journals/csur/LiuYFJHN23,DBLP:journals/corr/abs-2003-08271,DBLP:journals/corr/abs-2303-04226,DBLP:conf/ijcai/LiTZW21,DBLP:journals/corr/abs-2301-00234,DBLP:conf/acl/0009C23,llm_survey}. 
Compared with them, we focus on the techniques and methods for developing and applying LLMs for IR systems. In addition, we suggest reading the strategy report from the Chinese IR community~\cite{ir_perspective}, which discusses the opportunity and future directions of IR in the era of LLMs, and we think it is an excellent supplement to this survey.

The remaining part of this survey is organized as follows: Section~\ref{sec:bk} introduces the background for IR and LLMs. Section~\ref{sec:qr},~\ref{sec:ret},~\ref{sec:rank},~\ref{sec:reader} respectively review recent progress from the four perspectives of query rewriter, retriever, reranker, and reader, which are four key components of an IR system. Section~\ref{sec:agent} introduces recent studies of search agents. Then, Section~\ref{sec:future} discusses some potential directions in future research. Finally, we conclude the survey in Section~\ref{sec:conclu} by summarizing the major findings.

%% file: S2-background.tex
\section{Background}
\label{sec:bk}
\subsection{Information Retrieval}
Information retrieval (IR), as an essential branch of computer science, aims to efficiently retrieve information relevant to user queries from a large repository.  Generally, users interact with an IR system by submitting their queries in textual form. Subsequently, IR systems undertake the task of matching and ranking these user-supplied queries against an indexed database, thereby facilitating the retrieval of the most pertinent results.

The field of IR has witnessed significant advancement with the emergence of various models over time. One such early model is the Boolean model, which employs Boolean logic operators to combine query terms and retrieve documents that satisfy specific conditions~\cite{DBLP:books/mg/SaltonG83}. Based on the ``bag-of-words'' assumption, the vector space model~\cite{vector_space} represents documents and queries as vectors in term-based space. Relevance estimation is then performed by assessing the lexical similarity between the query and document vectors. The efficiency of this model is further improved through the effective organization of text content using the inverted index. Moving towards more sophisticated approaches, statistical language models have been introduced to estimate the likelihood of term occurrences and incorporate context information, leading to more accurate and context-aware retrieval~\cite{DBLP:conf/cikm/SongC99, DBLP:journals/arist/LiuC05}. In recent years, the neural IR paradigm has gained considerable attention in the research community~\cite{DBLP:conf/cikm/GuoFAC16, DBLP:journals/corr/MitraC17, DBLP:journals/corr/abs-2211-14876}. By harnessing the powerful representation capabilities of neural networks, this paradigm can capture semantic relationships between queries and documents, thereby significantly enhancing retrieval performance.

Researchers have identified several challenges with implications for the performance and effectiveness of IR systems, such as query ambiguity and retrieval efficiency. In light of these challenges, researchers have directed their attention toward crucial modules within the retrieval process, aiming to address specific issues and effectuate corresponding enhancements. The pivotal role of these modules in ameliorating the IR pipeline and elevating system performance cannot be overstated. In this survey, we focus on the following four modules, which have been greatly enhanced by LLMs.

\mypara{Query Rewriter} is an essential IR module that seeks to improve the precision and expressiveness of user queries. Positioned at the early stage of the IR pipeline, this module assumes the crucial role of refining or modifying the initial query to align more accurately with the user's information requirements. As an integral part of query rewriter, query expansion techniques, with pseudo relevance feedback being a prominent example, represent the mainstream approach to achieving query expression refinement. In addition to its utility in improving search effectiveness across general scenarios, the query rewriter finds application in diverse specialized retrieval contexts, such as personalized search and conversational search, thus further demonstrating its significance.

\mypara{Retriever}, as discussed here, is typically employed in the early stages of IR for document recall. The evolution of retrieval technologies reflects a constant pursuit of more effective and efficient methods to address the challenges posed by ever-growing text collections. In numerous experiments on IR systems over the years, the classical ``bag-of-words'' model BM25~\cite{DBLP:conf/trec/RobertsonWJHG94} has demonstrated its robust performance and high efficiency. In the wake of the neural IR paradigm's ascendancy, prevalent approaches have primarily revolved around projecting queries and documents into high-dimensional vector spaces, and subsequently computing their relevance scores through inner product calculations. This paradigmatic shift enables a more efficient understanding of query-document relationships, leveraging the power of vector representations to capture semantic similarities.

\mypara{Reranker}, as another crucial module in the retrieval pipeline, primarily focuses on fine-grained reordering of documents within the retrieved document set. Different from the retriever, which emphasizes the balance of efficiency and effectiveness, the reranker module places a greater emphasis on the quality of document ranking. In pursuit of enhancing the search result quality, researchers delve into more complex matching methods than the traditional vector inner product, thereby furnishing richer matching signals to the reranker. Moreover, the reranker facilitates the adoption of specialized ranking strategies tailored to meet distinct user requirements, such as personalized and diversified search results. By integrating domain-specific objectives, the reranker module can deliver tailored and purposeful search results, enhancing the overall user experience. 

\mypara{Reader} has evolved as a crucial module with the rapid development of LLM technologies. Its ability to comprehend real-time user intent and generate dynamic responses based on the retrieved text has revolutionized the presentation of IR results. In comparison to presenting a list of candidate documents, the reader module organizes answer texts more intuitively, simulating the natural way humans access information. To enhance the credibility of generated responses, the integration of references into generated responses has been an effective technique of the reader module.

Furthermore, researchers explore unifying the above modules to develop a novel LLM-driven search model known as the \textbf{Search Agent}. The search agent is distinguished by its simulation of an automated search and result understanding process, which furnishes users with accurate and readily comprehensible answers. WebGPT~\cite{webgpt} serves as a pioneering work in this category, which models the search process as a sequence of actions of an LLM-based agent within a search engine environment, autonomously accomplishing the whole search pipeline. By integrating the existing search stack, search agents have the potential to become a new paradigm in future IR.
 
\subsection{Large Language Models}
Language models (LMs) are designed to understand or generate human language by taking into account the contextual information from word sequences.
The evolution from \textbf{statistical language models} to \textbf{neural language models} makes it feasible to utilize LMs for representation learning beyond mere word sequence modeling.
\citet{elmo} first proposed to learn contextualized word representations through pre-training a bidirectional LSTM (biLSTM) network on large-scale corpora, followed by fine-tuning on specific downstream tasks. Similarly, \citet{bert} proposed to pre-train a Transformer~\cite{transformer} encoder with a specially designed Masked Language Modeling (MLM) task and Next Sentence Prediction (NSP) task on large corpora. These studies initiated a new era of \textbf{pre-trained language models} (PLMs), with the ``pre-training then fine-tuning'' paradigm emerging as the prevailing learning approach. Along this line, numerous generative PLMs (\eg, GPT-2~\cite{gpt-2}, BART~\cite{bart}, and T5~\cite{t5}) have been developed for text generation problems including summarization, machine translation, and dialogue generation. Recently, researchers have observed that increasing the scale of PLMs (\eg, model size or data amount) can consistently improve their performance on downstream tasks (a phenomenon commonly referred to as the \textit{scaling law}~\cite{scalinglaw_openai,scalinglaw_deepmind}). Moreover, large-sized PLMs exhibit promising abilities (termed \textit{emergent abilities}~\cite{emergent_ability}) in addressing complex tasks, which are not evident in their smaller counterparts. Therefore, the research community refers to these large-sized PLMs \textbf{as large language models} (LLMs).



Owing to their vast number of parameters, fine-tuning LLMs for specific tasks, such as IR, is often deemed impractical. Consequently, two prevailing methods for applying LLMs have been established: in-context learning (ICL) and parameter-efficient fine-tuning. ICL is one of the emergent abilities of LLMs~\cite{gpt-3} empowering them to comprehend and furnish answers based on the provided input context, rather than relying merely on their pre-training knowledge. This method requires only the formulation of the task description and demonstrations in natural language, which are then fed as input to the LLM. Notably, parameter tuning is not required for ICL. Additionally, the efficacy of ICL can be further augmented through the adoption of chain-of-thought prompting, involving multiple demonstrations (describe the chain of thought examples) to guide the model's reasoning process. ICL is the most commonly used method for applying LLMs to IR. Parameter-efficient fine-tuning~\cite{lora,DBLP:conf/acl/LiL20,DBLP:conf/emnlp/LesterAC21} aims to reduce the number of trainable parameters while maintaining satisfactory performance. LoRA~\cite{lora}, for example, has been widely applied to open-source LLMs (\eg, LLaMA and BLOOM) for this purpose. Recently, QLoRA~\cite{qlora} has been proposed to further reduce memory usage by leveraging a frozen 4-bit quantized LLM for gradient computation. Despite the exploration of parameter-efficient fine-tuning for various NLP tasks, its implementation in IR tasks remains relatively limited, representing a potential avenue for future research.

Recently, research has focused on enhancing LLM capabilities by improving their reasoning and inference-time processes. \textbf{Large reasoning models} (LRMs) are an evolution of LLMs specifically designed to excel at complex logical tasks such as mathematics and coding. These models often employ advanced training methodologies, including reinforcement learning, to develop sophisticated self-reflection and planning mechanisms. They are designed to generate a detailed ``thinking process'' before providing a final answer, which has shown improved performance on reasoning benchmarks. This focus on improving the reasoning process is highly relevant to test-time scaling, a paradigm that allocates additional computational resources during inference to improve model performance without increasing model size during pre-training. This can involve generating multiple outputs in parallel and selecting the best one, or employing sequential methods where the model iteratively refines its own output. By achieving promising results on expert-level human challenges, these LRMs provide a new paradigm for IR systems.

%% file: S3-query-rewriter.tex
\begin{figure}[t]
    \centering
    \includegraphics[width=\linewidth]{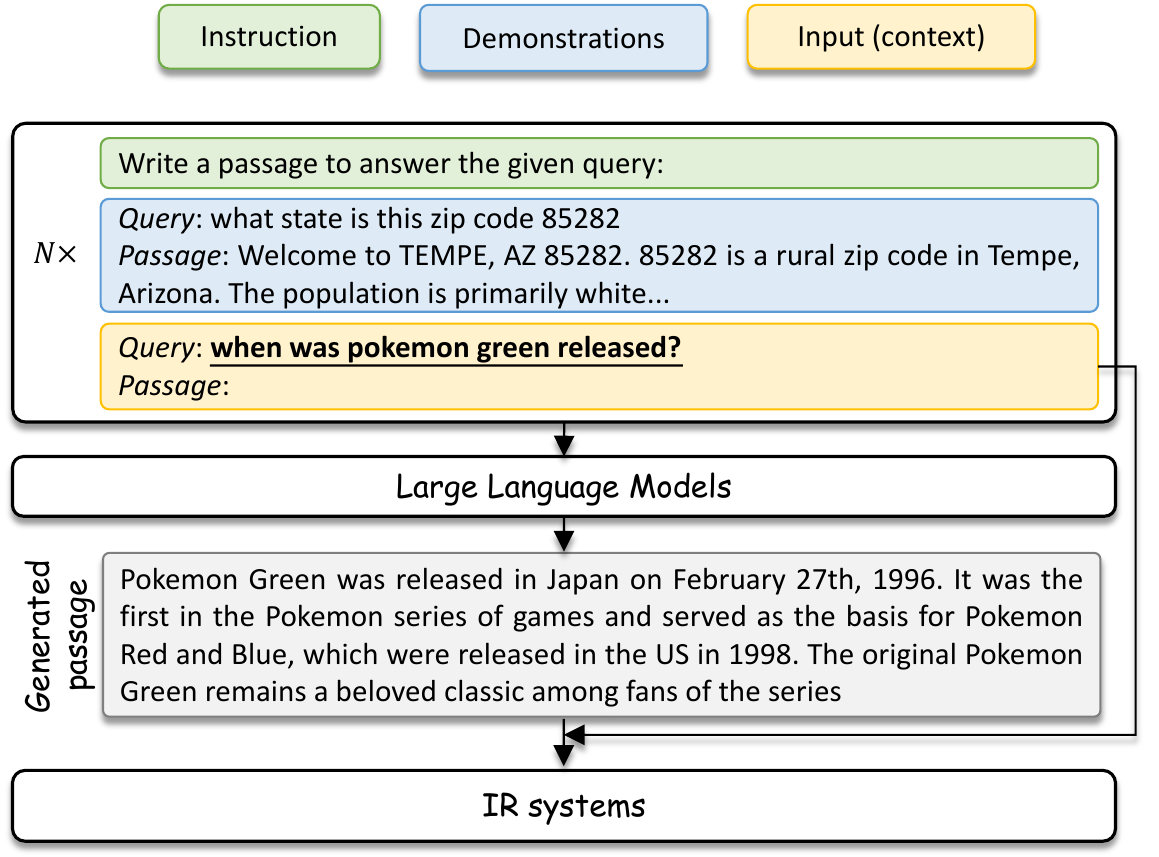}
    \caption{An example of LLM-based query rewriter for ad-hoc search. The example is cited from the Query2Doc paper~\cite{wang2023query2doc}. LLMs are used to generate a passage to supplement the original query, where $N=0$ and $N>0$ correspond to zero-shot and few-shot scenarios.}
    \label{fig:query_rewrite_adhoc}
\end{figure}

\section{Query Rewriter}
\label{sec:qr}

Query rewriter, functioning as an essential preprocessing component for search engines, increases the accuracy of retrieval systems through the refinement of initial queries~\cite{DBLP:journals/ipm/AzadD19}. This mechanism, also known as query expansion or reformulation, holds a pivotal position in search engine operations. In the context of ad-hoc retrieval, the design of a query rewriter aims to mitigate the vocabulary mismatch problem by enriching original queries with semantically related terms. As conversational search evolves, query rewriters have evolved to interpret user intent and previous dialogues, thereby enabling context-sensitive queries. In this survey, the term ``query rewriter'' is used to refer to any technique that improves retrieval performance through query modification.

Traditional query rewriting strategies primarily include techniques such as utilizing lexical knowledge bases~\cite{DBLP:journals/jasis/PeatW91,fellbaum1998wordnet,DBLP:journals/sigir/Fox80,DBLP:journals/jocch/ZoharLSD13,DBLP:journals/tois/GauchWR99} and pseudo-relevance feedback~\cite{li2007improving,DBLP:conf/ictir/XiongC15,DBLP:journals/nca/SinghS17}. However, these methods are limited due to the inadequate capabilities of knowledge models and the presence of noisy signals from coarse matching between the query and the top-$k$ retrieved documents. LLMs, pretrained on vast datasets, demonstrate considerable advancements in the breadth of knowledge and language understanding, positioning them as an excellent resource for query rewriting tasks. In the subsequent sections, we provide a comprehensive review of recent research that applies LLMs to query rewriting.

\begin{table}[t]
    \centering
    \caption{Overview of existing LLM-based query rewriting methods. ``Knowledge'' denotes the source of information the method employs for query rewriting. ``SFT'' and ``RL'' denotes supervised fine-tuning and reinforcement learning, respectively. ``$\mathcal{Q}$'', ``$\mathcal{K}$'', and ``$\mathcal{A}$'' refer to question, keyword, and answer-incorporated passages, respectively.}
    \setlength{\tabcolsep}{.9mm}{
    \begin{tabular}{lllcl}
    \toprule
            \textbf{Scenario} & \textbf{Knowledge} & \textbf{Approach} & \textbf{Format} & \textbf{Method} \\
    \midrule
        \multirow{15}{*}{Ad-hoc} & LLMs & {Prompting} & $\mathcal{A}$ & HyDE~\cite{gao2022precise} \\ 
              & LLM & Prompting & $\mathcal{A}$ & \citet{jagerman2023query} \\
              & LLM & Prompting & $\mathcal{A}$ &Query2Doc~\cite{wang2023query2doc} \\ 
              & LLM & Prompting & $\mathcal{A}$ & \citet{craft_the_path} \\
              & LLM & Prompting & $\mathcal{Q}$ &\citet{sigirAlaofi} \\ 
              & LLM & Prompting & $\mathcal{K}$ &\citet{keyword_and_refine}
              \\
              & LLM & RL & $\mathcal{K}$&\citet{ma2023query} 
              \\
              & LLM & SFT \& RL & $\mathcal{K}$&BEQUE~\cite{BEQUE} \\
              & LLM + Corpora & {Prompting}&$\mathcal{K}$ & GRF+PRF~\cite{GRF+PRF} \\ 
              & LLM + Corpora & Prompting & $\mathcal{A}$ &GRM~\cite{GRM} \\ 
              & LLM + Corpora & Prompting & $\mathcal{A}$ &InteR~\cite{RoundInteraction} \\ 
              & LLM + Corpora & Prompting & $\mathcal{A}$ & LameR~\cite{LameR} \\ 
              & LLM + Corpora & Prompting & $\mathcal{A}$ &CSQE~\cite{CSQE} \\
              & LLM + Corpora & Prompting &$\mathcal{Q}$ & CAR~\cite{CAR} \\
              & LLM + Corpora & SFT \& RL  & $\mathcal{Q}$ & RaFe~\cite{RaFe} \\
        \midrule
        \multirow{3}{*}{\makecell[c]{Conver- \\sational}} & LLM & Prompting & $\mathcal{Q}$ &LLMCS~\cite{kelong_conversational}\\
              & LLM & Prompting & $\mathcal{Q}$ &CONVERSER~\cite{Converser}\\ 
              & LLM & Prompting & $\mathcal{Q}$ & \citet{ye2023enhancing} \\ 
    \bottomrule
    \end{tabular}
    }
    \label{tab:qr_classification}
\end{table}

\subsection{Rewriting Scenarios}

\begin{figure}[t]
    \centering
    \includegraphics[width=\linewidth]{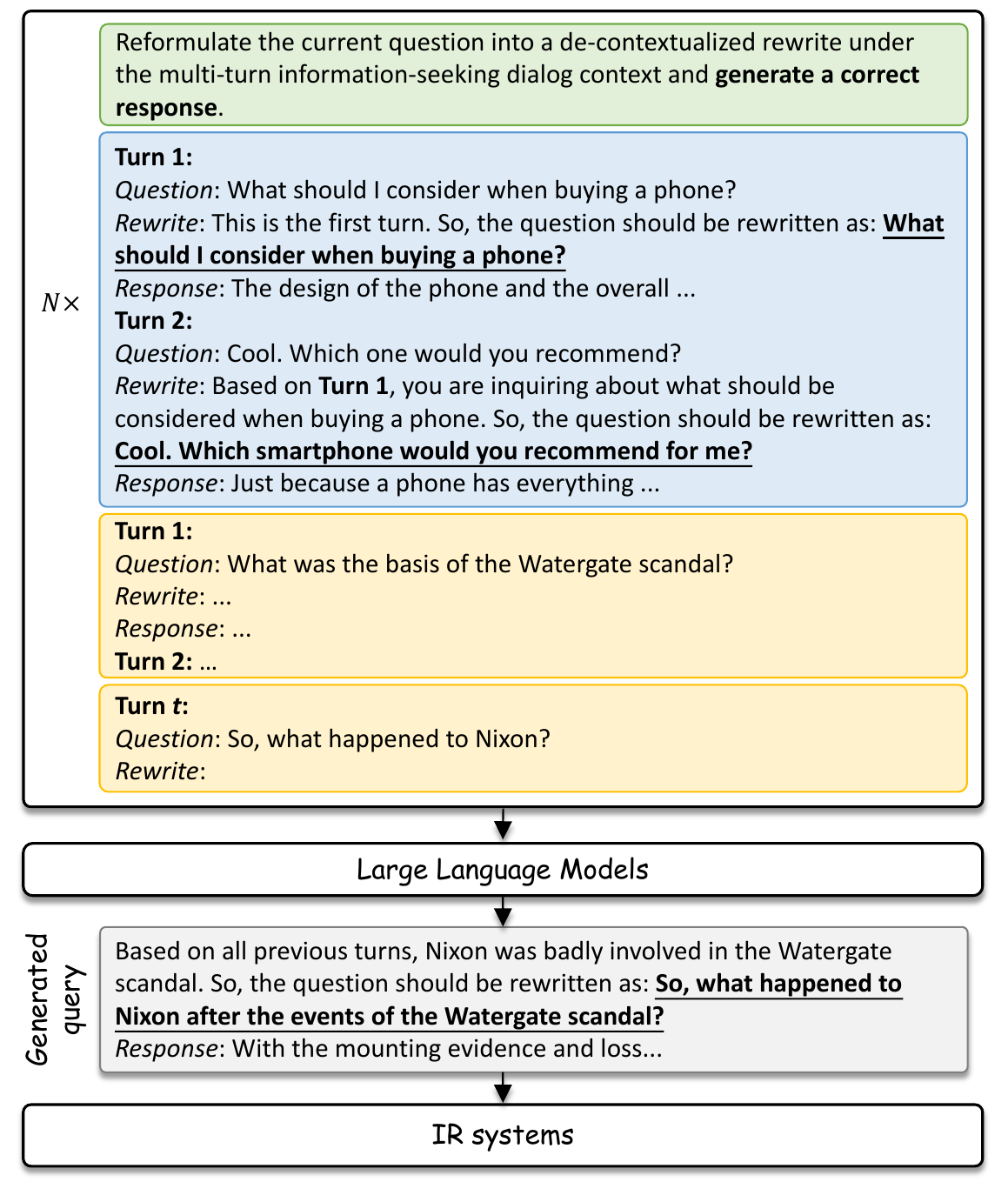}
    \caption{An example of LLM-based query rewriter for conversational search (cited from LLMCS~\cite{kelong_conversational}) . An LLM is used to generate a query and system response based on the demonstrations and previous search context. $N=0$ and $N>0$ correspond to zero-shot and few-shot scenarios respectively.}
    \label{fig:query_rewrite_conv}
\end{figure}

In the realm of IR, a query rewriter is primarily designed to serve two distinct scenarios: ad-hoc retrieval and conversational search.
Ad-hoc retrieval aims to bridge the semantic gap between a user's query and the potential documents. LLMs, with their extensive inherent knowledge, have proven effective in replacing traditional lexical knowledge databases~\cite{DBLP:journals/jasis/PeatW91,fellbaum1998wordnet,DBLP:journals/sigir/Fox80,DBLP:journals/jocch/ZoharLSD13,DBLP:journals/tois/GauchWR99}. 

For conversational search, query rewriters aim to refine a query within a conversation's context, transforming it into isolated queries based on historical dialogues. A crucial requirement for conversational query rewriters is to address coreference resolution. Traditional query rewriting methods, trained on limited data, have shown suboptimal performance, as conversational search sessions tend to be diverse and long-tailed~\cite{DBLP:conf/emnlp/MaoDQMCC22,DBLP:conf/icml/DaiCZARGG22}. This is particularly the case in more complex conversational search sessions. However, LLMs, with their robust context understanding capabilities, have demonstrated significant advantages in conversational query rewriting~\cite{kelong_conversational,Converser}.

Beyond traditional retrieval scenarios, query rewriting is also widely used in a variety of practical domains. In the context of agents, effectively identifying the most relevant tools for a given task becomes a key bottleneck as the toolset size grows, hindering reliable tool utilization. To address this, current study~\cite{chen-etal-2024-invoke} propose generating a diverse set of synthetic queries that comprehensively cover different aspects of the query space associated with each tool document during the tool indexing phase. On the other hand, in clinical terminology normalization, recent study~\cite{fan-etal-2024-rrnorm} also leverage large language models to decompose and reconstruct complex diagnostic mentions, improving the mapping to standard terms through a "retrieve-and-rank" framework to enhance overall performance.

By leveraging the capabilities of LLMs, researchers have been able to generate a variety of formats for rewritten queries, such as questions~\cite{sigirAlaofi,ma2023query,An_Interactive_Query_Generation,ye2023enhancing,RaFe,kelong_conversational,Converser, wilson2025contextualizingsearchqueriesincontext}, answer-incorporated passages~\cite{wang2023query2doc,craft_the_path,GRM,RoundInteraction,LameR,gao2022precise,jagerman2023query,CAR, KELLER}, and keywords~\cite{GRF+PRF,keywords,BEQUE,ma2023query}. Comprehensive details for each format are available in the following part.

\subsection{Formats of Rewritten Queries}
The intended format for rewritten queries can vary widely based on the specific needs and the downstream retrieval system. The ultimate goal is to improve the effectiveness of IR. Typically, the formats include questions, keywords, and answer-incorporated passages. 

\subsubsection{Questions}
Rewriting original queries into similar form questions are a natural idea of query rewriting~\cite{sigirAlaofi,ye2023enhancing,RaFe}. Query rewriters modify original queries to new questions to make it more precise, understandable, and aligned with the user's actual search intent. This can involve rephrasing, expanding, or simplifying the query. Recent research~\cite{sigirAlaofi} has demonstrated the potential of using LLMs to generate query variants. Although these variants cannot cover the full range of human-generated ones, they do produce highly similar sets of relevant documents.

\subsubsection{Keywords}
Keywords serve as a high-level abstraction of the concepts contained within a query. Rewriting queries into keywords proves particularly effective when the downstream retriever is a sparse retriever. With specific instructions, LLMs can produce high-quality keywords or concepts for query rewriting~\cite{ma2023query, jagerman2023query, BEQUE, GRF+PRF}. For example, BEQUE~\cite{BEQUE} formulates new queries as keywords for effective product searches, and \citet{keyword_and_refine} introduce a two-round query rewriting process, which first generates a set of high-quality seed keywords, then utilizes these keywords to enhance the query.

\subsubsection{Answer-incorporated Passages}
The semantic gap between short-form queries and long-form documents has been a persistent challenge. The advent of LLMs with their inherent question-answering capabilities has introduced a novel approach to query rewriting. This approach involves initially utilizing LLMs to generate comprehensive answers to the given queries. These detailed answers are then employed to retrieve relevant passages from the corpus, thereby effectively bridging the semantic divide between short queries and long candidate documents. The prompt employed for this mechanism is typically structured as follows: ``Given a question {query} and its potential answer passages {passages}, compose a passage that provides an answer to the question''~\cite{LameR,RoundInteraction}. This approach enables a more nuanced and contextually relevant retrieval of information, enhancing the overall effectiveness of the query rewriting process~\cite{wang2023query2doc,craft_the_path,GRM,RoundInteraction,LameR,gao2022precise,jagerman2023query}.

\subsection{Approaches}
The utilization of LLMs in query rewriting can be categorized into three primary methodologies: \textit{prompting}, \textit{supervised fine-tuning}, and \textit{reinforcement learning}. The {prompting} approaches employ specific prompts to guide the LLM's output, providing flexibility and interpretability. The {supervised fine-tuning} techniques adapt pre-trained LLMs to the specific task of query rewriting. However, the scarcity of training data for query rewriting often poses a challenge. To address this issue, {reinforcement learning} methods utilize feedback from downstream applications, thereby improving the performance of query rewriters. In the following section, we will introduce these three methods in detail.

\subsubsection{Prompting}
\begin{table}[t]
    \centering
    \caption{Examples of different prompting methods in query rewriter.}
    \begin{tabular}{p{.18\linewidth}p{.72\linewidth}}
    \toprule
    \textbf{Methods} & \textbf{Prompts} \\
    \midrule
    \multicolumn{2}{c}{\textit{Zero-shot}} \\
    \midrule
    {HyDE~\cite{gao2022precise}} & Please write a passage to answer the question. Question: \{\#Question\} Passage: \\
    {LameR~\cite{LameR}} & Give a question \{\#Question\} and its possible answering passages: A. \{\#Passage 1\} B. \{\#Passage 2\} C. \{\#Passage 3\} ... Please write a correct answering passage.\\
    \midrule
    \multicolumn{2}{c}{\textit{Few-shot}} \\
    \midrule
    {Query2Doc~\cite{wang2023query2doc}} & Write a passage that answers the given query: \\
     & Query: \{\#Query 1\} \\
     & Passage: \{\#Passage 1\} \\
     & ... \\
     & Query: \{\#Query\} \\
     & Passage: \\
    \midrule
    \multicolumn{2}{c}{\textit{Chain-of-Thought}} \\
    \midrule
    {CoT~\cite{jagerman2023query}} & Answer the following query based on the context: \\
    & Context: \{\#PRF doc 1\} \{\#PRF doc 2\} \{\#PRF doc 3\} \\
    & Query: \{\#Query\} \\
    & Give the rationale before answering \\
    \bottomrule
    \end{tabular}
    \label{tab:prompting_examples}
\end{table}

Prompting in LLMs refers to the technique of providing a specific instruction or context to guide the model's generation of text. The prompt serves as a conditioning signal and influences the language generation process of the model. Existing prompting strategies can be roughly categorized into three groups: zero-shot prompting, few-shot prompting, and chain-of-thought (CoT) prompting~\cite{cot}. 

$\bullet$ \textit{Zero-shot prompting.} Zero-shot prompting involves instructing the model to generate texts on a specific topic without any prior exposure to training examples in that domain or topic. The model relies on its pre-existing knowledge and language understanding to generate coherent and contextually relevant expanded terms for original queries. Experiments show that zero-shot prompting is a simple yet effective method for query rewriter~\cite{sigirAlaofi,wenhao_generate,GRF+PRF,LameR,RoundInteraction,jagerman2023query}.

$\bullet$ \textit{Few-shot prompting.} Few-shot prompting, also known as in-context learning, involves providing the model with a limited set of examples or demonstrations related to the desired task or domain~\cite{sigirAlaofi,wenhao_generate,jagerman2023query,wang2023query2doc}. These examples serve as a form of explicit instruction, allowing the model to adapt its language generation to specific tasks or domains. Query2Doc~\cite{wang2023query2doc} prompts LLMs to write a document that answers the query with some demo query-document pairs provided by the ranking dataset, such as MSMARCO~\cite{MSMARCO} and NQ~\cite{nq}. This work experiments with a single prompt. To further study the impact of different prompt designing, recent works~\cite{jagerman2023query} have explored eight different prompts, such as prompting LLMs to generate query expansion terms instead of entire pseudo documents and CoT prompting. Some illustrative prompts are shown in Table~\ref{tab:prompting_examples}. The experiments validate that Query2Doc is more effective than many other prompt-based methods.

$\bullet$ \textit{Chain-of-thought prompting.} CoT prompting~\cite{cot} is a strategy that involves iterative prompting, where the model is provided with a sequence of instructions or partial outputs~\cite{sigirAlaofi,jagerman2023query,wu2025cotkrchainofthoughtenhancedknowledge,baek2024craftingpathrobustquery}. 
In conversational search, the process of query rewriting is multi-turn, which means queries should be refined step-by-step with the interaction between search engines and users. This process naturally coincides with the CoT process. As shown in Figure~\ref{fig:query_rewrite_conv}, users can conduct the CoT process by adding some instructions during each turn, such as ``Based on all previous turns, xxx''. While in ad-hoc search, there is only one round in query rewriting, so CoT could only be accomplished in a simple and coarse way. For example, as shown in Table~\ref{tab:prompting_examples}, researchers add ``Give the rationale before answering'' in the instructions to prompt LLMs to think deeply~\cite{jagerman2023query}.

\subsubsection{Supervised Fine-tuning}
Prompting methods directly leverage LLMs' strong capabilities to expand or rewrite queries. Though prompting method is effective, LLMs are not naturally designed for query rewriting task. To further tailor LLMs for this task, supervised fine-tuning (SFT) has emerged as a promising approach. A crucial aspect of this methodology is the creation of an appropriate training dataset. The process of gathering this dataset varies significantly depending on the application scenario.

In the context of e-commerce search, a wealth of supervised training data for query rewriting is naturally available. This data, sourced from the previous-generation rewriting policies of the e-commerce system, significantly simplifies the construction of the SFT dataset~\cite{BEQUE}.

Conversely, in an ad-hoc retrieval scenario, the acquisition of query rewrite training data is often a challenge. To address this issue, researchers usually employ implicit feedback and reinforcement learning to train the query rewriter.

\subsubsection{Reinforcement Learning} 
Query rewriters typically serve as intermediaries for retrieval systems, and as such, they lack a dedicated or independent loss function for optimization. In this context, reinforcement learning (RL) presents an alternative training paradigm. The query rewriter can receive feedback signals from donwstream components, such as ranking models~\cite{RaFe} or LLM readers~\cite{ma2023query}. For instance, ranking scores can be utilized to construct good-bad pairs for direct preference optimization~\cite{dpo} training. Similarly, \citet{ma2023query} propose to generate answers from LLMs and then uses the results of a QA evaluation as training signals. Another approach, BEQUE~\cite{BEQUE}, introduces an offline feedback system that assigns a quality score to each query based on the set of products it retrieves. 

Recently, inspired by the rule-based reward system proposed by DeepSeek-R1~\cite{r1}, some studies have explored using retrieval metrics as the reward to optimize query generators. For example, DeepRetrieval~\cite{jiang2025deepretrievalhackingrealsearch} proposes a RL approach that trains LLMs for query generation by using retrieval metrics as rewards without the need for supervised data. This method reinforces the query generator to produce queries that maximize retrieval performance. These RL mechanisms align the objective of query rewriters more closely with the goals of downstream tasks, thereby enhancing the overall performance of the system.

\subsection{Limitations}
Despite the potential of LLMs in query rewriting, they still suffer from several limitations. We discuss two primary challenges that arise with their use in this context.

\subsubsection{Concept Drifts}
One significant issue is the introduction of unrelated information, or concept drift, which may occur due to the LLM's vast knowledge base and propensity to generate detailed yet sometimes redundant content. While this can potentially enrich the query, it may also lead to irrelevant or off-target results. This issue has been reported in various studies~\cite{An_Interactive_Query_Generation,CAR,BEQUE}. These works emphasize the necessity for a balanced approach to LLM-based query rewriting, maintaining the core and focus of the original query while utilizing the LLM's capabilities to enhance and clarify it. This balance is critical for effective search and IR applications.

\subsubsection{Correlation between Retrieval Performance and Expansion Effects}
A recent comprehensive study~\cite{DBLP:journals/corr/when_do_generative} has investigated various expansion techniques and downstream ranking models, revealing a significant negative correlation between retrieval performance and expansion benefits. Specifically, expansion tends to improve the scores of weaker models but adversely affects stronger ones. This finding suggests a strategic approach: only using expansions with weaker models or when the target dataset significantly varies in format from the training corpus. In other situations, it may be more beneficial to refrain from expansions to preserve the clarity of the relevance signal.

%% file: S4-retriever.tex
\section{Retriever}
\label{sec:ret}
In an IR system, the retriever serves as the first-pass document filter to collect broadly relevant documents for user queries. Given the enormous amounts of documents in an IR system, the retriever's efficiency in locating relevant documents is essential for maintaining search engine performance. Meanwhile, a high recall is also important for the retriever, as the retrieved documents are then fed into the ranker to generate final results for users, which determines the ranking quality of search engines.

In recent years, retrieval models have shifted from relying on statistic algorithms~\cite{DBLP:conf/trec/RobertsonWJHG94} to neural models~\cite{ance,dpr}. The latter approaches exhibit superior semantic capability and excel at understanding complicated user intent. The success of neural retrievers relies on two key factors: \emph{data} and \emph{model}. From the data perspective, a large amount of high-quality training data is essential. This enables retrievers to acquire comprehensive knowledge and accurate matching patterns. Furthermore, the intrinsic quality of search data, \ie, issued queries and document corpus, significantly influences retrieval performance. From the model perspective, a strongly representational neural architecture allows retrievers to effectively store and apply knowledge obtained from the training data. 

Unfortunately, there are some long-term challenges that hinder the advancement of retrieval models. First, user queries are usually short and ambiguous, making it difficult to precisely understand the user's search intents for retrievers. Second, documents typically contain lengthy content and substantial noise, posing challenges in encoding long documents and extracting relevant information for retrieval models. Additionally, the collection of human-annotated relevance labels is time-consuming and costly. It restricts the retrievers' knowledge boundaries and their ability to generalize across different application domains. Moreover, existing model architectures, primarily built on BERT~\cite{bert}, exhibit inherent limitations, thereby constraining the performance potential of retrievers.
Recently, LLMs have exhibited extraordinary abilities in language understanding, text generation, and reasoning. This has motivated researchers to use these abilities to tackle the aforementioned challenges and aid in developing superior retrieval models. Roughly, these studies can be categorized into two groups: (1) leveraging LLMs to generate search data, and (2) employing LLMs to enhance model architecture. 

\subsection{Leveraging LLMs to Generate Search Data}
In light of the quality and quantity of search data, there are two prevalent perspectives on how to improve retrieval performance via LLMs. The first perspective revolves around search data refinement methods, which concentrate on reformulating input queries to precisely present user intents. The second perspective involves training data augmentation methods, which leverage LLMs' generation ability to enlarge the training data for dense retrieval models, particularly in zero- or few-shot scenarios.



\subsubsection{Search Data Refinement}
Typically, input queries consist of short sentences or keyword-based phrases that may be ambiguous and contain multiple possible user intents. Accurately determining the specific user intent is essential in such cases. Moreover, documents usually contain redundant or noisy information, which poses a challenge for retrievers to extract relevance signals between queries and documents.
Leveraging the strong text understanding and generation capabilities of LLMs offers a promising solution to these challenges. As yet, research efforts in this domain primarily concentrate on employing LLMs as query rewriters, aiming to refine input queries for more precise expressions of the user's search intent. Section~\ref{sec:qr} has provided a comprehensive overview of these studies, so this section refrains from further elaboration. 
In addition to query rewriter, an intriguing avenue for exploration involves using LLMs to enhance the effectiveness of retrieval by refining lengthy documents. This intriguing area remains open for further investigation and advancement.

\begin{figure*}
    \centering
    \includegraphics[width=1\linewidth]{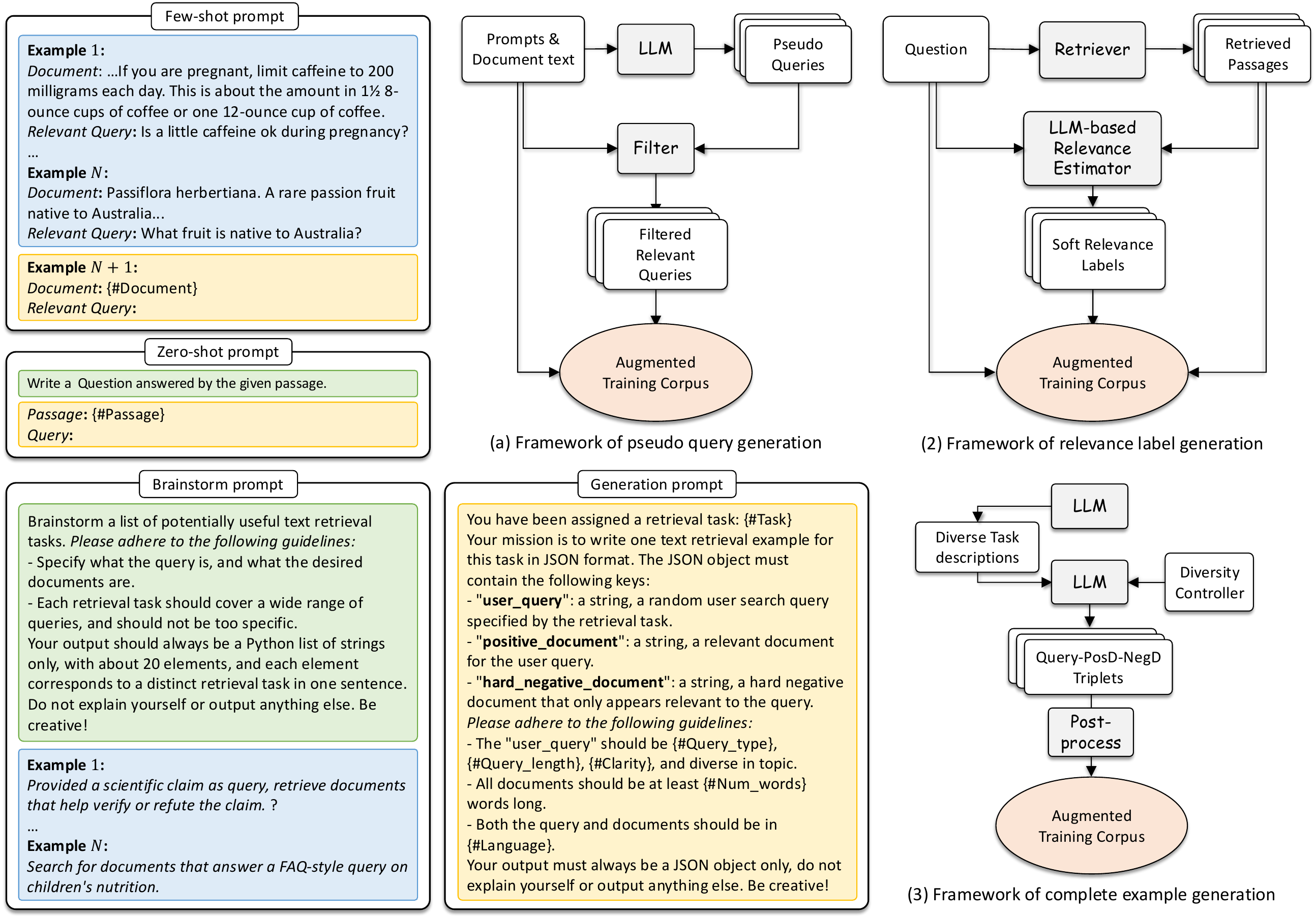}
    \caption{Three typical frameworks for LLM-based data augmentation in the retrieval task (right), along with their prompt
examples (left). Note that the methods of relevance label generation do not treat questions as inputs but regard their
generation probabilities conditioned on the retrieved passages as soft relevance labels.}
    \label{fig:aug-frame}
\end{figure*}

\begin{table*}[t]
    \centering
    \caption{The comparison of existing data augmentation methods powered by LLMs for training retrieval models.}
    \setlength{\tabcolsep}{1.7mm}{
    \begin{tabular}{lccccc}
    \toprule
    \textbf{Methods} & \textbf{\# Examples} & \textbf{Generator} & \textbf{Synthetic Data} & \textbf{Filter Method}  & \textbf{LLMs' tuning} \\
    \midrule
    InPairs~\cite{InPairs}           & 3 & Curie & {Relevant query} & {Generation probability} & Fixed \\
    \citet{Bottlenecked_Query_Gen}   & 0-2 & Alpaca-LLaMA \& tk-Instruct & Relevant query & - & Fixed \\
    InPairs-v2~\cite{InPairs-v2}     & 3 & GPT-J & {Relevant query} & \makecell{Relevance score from\\fine-tuned monoT5-3B} & Fixed \\
    PROMPTAGATOR~\cite{PROMPTAGATOR} & 0-8 & FLAN & Relevant query & Round-trip filtering & Fixed \\
    TQGen~\cite{TQGen}               & 0 & T0 & Relevant query & Generation probability & Fixed  \\
    UDAPDR~\cite{UDAPDR}             &  0-3 & GPT3 \& FLAN-T5-XXL & Relevant query & Round-trip filtering & Fixed \\
    SPTAR~\cite{SoftPrompt}          & 1-2 & LLaMA-7B \& Vicuna-7B & Relevant query & BM25 filtering & Soft Prompt tuning \\
    Gecko~\cite{Gecko}               & few-shot & Unkown & Relevant query & - & Fixed \\
    ART~\cite{ART}                   & 0 & T5-XL \& T5-XXL & Soft relevance labels & - & Fixed \\
    \citet{GENQD}                    & 2 & GPT-4 & Q-PosD-NegD triplet & - & fixed \\
    \bottomrule
    \end{tabular}
    }
    \label{tab:retrieve_dataaug_comparison}
\end{table*}

\subsubsection{Training Data Augmentation}
Due to the expensive economic and time costs of human-annotated labels, a common problem in training neural retrieval models is the lack of training data. Fortunately, the excellent capability of LLMs in text generation offers a potential solution. A key research focus lies in devising strategies to leverage LLMs' capabilities to generate pseudo-relevant signals and augment the training dataset for the retrieval task.

\mypara{Why do we need data augmentation?}
Previous studies of neural retrieval models focused on supervised learning, namely training retrieval models using labeled data from specific domains. For example, MS MARCO~\cite{MSMARCO} provides a vast repository, containing a million passages, more than 200,000 documents, and 100,000 queries with human-annotated relevance labels, which has greatly facilitated the development of supervised retrieval models. However, this paradigm inherently constrains the retriever's generalization ability for out-of-distribution data from other domains. The application spectrum of retrieval models varies from natural question-answering to biomedical IR, and it is expensive to annotate relevance labels for data from different domains. As a result, there is an emerging need for zero-shot and few-shot learning models to address this problem~\cite{BEIR}. A common practice to improve the models' effectiveness in a target domain without adequate label signals is through data augmentation.

\mypara{How to apply LLMs for data augmentation?}
In the scenario of IR, it is easy to collect numerous documents. However, the challenging and costly task lies in gathering real user queries and labeling the relevant documents accordingly. Considering the strong text generation capability of LLMs, many researchers~\cite{InPairs,InPairs-v2} suggest using LLM-driven processes to create pseudo queries or relevance labels based on existing collections. These approaches facilitate the construction of relevant query-document pairs, enlarging the training data for retrieval models. According to the type of generated data, there are three mainstream approaches that complement the LLM-based data augmentation for retrieval models, \ie, pseudo query generation, relevance label generation, and complete example generation. Their frameworks are visualized in Figure~\ref{fig:aug-frame}. 

$\bullet$ \textit{Pseudo query generation.} Given the abundance of documents, a straightforward idea is to use LLMs for generating their corresponding pseudo queries. One such illustration is presented by inPairs~\cite{InPairs}, which leverages the in-context learning capability of GPT-3. This method employs a collection of query-document pairs as demonstrations. These pairs are combined with a document and presented as input to GPT-3, which subsequently generates possible relevant queries for the given document. By combining the same demonstration with various documents, it is easy to create a vast pool of synthetic training samples and support the fine-tuning of retrievers on specific target domains. 
To enhance the diversity of generated examples, Gecko~\cite{Gecko} prompts LLMs to first generate a task description, and then generate pseudo queries according to the task. Considering the false negative problems, it further develops an LLM-based positive and negative mining strategy to discover potential relevant and hard negative documents from the corpus for generated queries, which significantly enhance the retrieval performance.  
Recent studies~\cite{Bottlenecked_Query_Gen} have also leveraged open-sourced LLMs, such as Alpaca-LLaMA and tk-Instruct, to produce sufficient pseudo queries and applied curriculum learning to pre-train dense retrievers. To enhance the reliability of these synthetic samples, a fine-tuned model (\eg, a monoT5-3B model fine-tuned on MSMARCO~\cite{InPairs-v2}) is employed to filter the generated queries. Only the top pairs with the highest estimated relevance scores are kept for training. This ``generating-then-filtering'' paradigm can be conducted iteratively in a round-trip filtering manner, \ie, by first fine-tuning a retriever on the generated samples and then filtering the generated samples using this retriever. Repeating these EM-like steps until convergence can produce high-quality training sets~\cite{PROMPTAGATOR}. Furthermore, by adjusting the prompt given to LLMs, they can generate queries of different types. This capability allows for a more accurate simulation of real queries with various patterns~\cite{TQGen}.

In practice, it is costly to generate a substantial number of pseudo queries through LLMs. Balancing the generation costs and the quality of generated samples has become an urgent problem. To tackle this, UDAPDR~\cite{UDAPDR} is proposed, which first produces a limited set of synthetic queries using LLMs for the target domain. These high-quality examples are subsequently used as prompts for a smaller model to generate a large number of queries, thereby constructing the training set for that specific domain. It is worth noting that the aforementioned studies primarily rely on fixed LLMs with frozen parameters. Empirically, optimizing LLMs' parameters can significantly improve their performance on downstream tasks. Unfortunately, this pursuit is impeded by the prohibitively high demand for computational resources. To overcome this obstacle, SPTAR~\cite{SoftPrompt} introduces a soft prompt tuning technique that only optimizes the prompts' embedding layer during the training process. This approach allows LLMs to better adapt to the task of generating pseudo-queries, striking a favorable balance between training cost and generation quality.

In addition to the above studies, pseudo query generation methods are also introduced in other application scenarios, such as conversational dense retrieval~\cite{Converser} and multilingual dense retrieval~\cite{Multiling_DR}.

$\bullet$ \textit{Relevance label generation.}
In some downstream tasks of retrieval, such as question-answering, the collection of questions is also sufficient. However, the relevance labels connecting these questions with the passages of supporting evidence are very limited. In this context, leveraging the capability of LLMs for relevance label generation is a promising approach that can augment the training corpus for retrievers. A recent method, ART~\cite{ART}, exemplifies this approach. It first retrieves the top-relevant passages for each question. Then, it employs an LLM to produce the generation probabilities of the question conditioned on these top passages. After a normalization process, these probabilities serve as soft relevance labels for the training of the retriever.

$\bullet$ \textit{Complete example generation.}
Recent study~\cite{GENQD} further investigates approaches to directly utilize LLMs to generate synthetic queries and documents, hence providing tremendous diverse training examples across varied tasks and languages. 
This work proposes a two-stage generation pipeline, where the first stage prompts the LLM to brainstorm various retrieval tasks, and then the second stage generates corresponding ``(query, positive document, negative document)'' triplets to build synthetic training data. Researchers further control the length, languages, and semantic relationships of queries and documents to produce diverse training samples. The generated triplets, after post-processing, \eg, de-duplication and JSON-format filtering, are used as training examples for optimizing dense retrievers.

Additionally, to highlight the similarities and differences among the corresponding methods, we present a comparative result in Table~\ref{tab:retrieve_dataaug_comparison}. It compares the aforementioned methods from various perspectives, including the number of examples, the generator employed, the type of synthetic data produced, the method applied to filter synthetic data, and whether LLMs are fine-tuned. This table serves to facilitate a clearer understanding of the landscape of these methods.

\subsection{Leveraging LLMs as Retrievers' Backbone}

Thanks to the superior text representation capability, LLMs excel at comprehending the underlying semantics of queries and documents. Therefore, it becomes increasingly popular to apply such large-scale models as the backbone of text retrievers, leading to substantial improvements over the conventional methods based on smaller-sized models~\cite{bert}. 





\subsubsection{Dense Retriever}
The application of LLMs brings about two major impacts on dense retrieval. On one hand, it advances the ongoing progress of the existing methods, making substantial improvements in terms of both in-domain accuracy and out-of-domain generalizability. On the other hand, it extends the boundaries of current methods, introducing new capabilities such as instruction following and in-context learning. 


\mypara{Improved Existing Capacities.}
Dense retrieval needs to fine-tune pre-trained text encoders such that queries and documents can be transformed into semantic-rich embeddings. Therefore, the downstream retrieval performance can benefit from the utilization of stronger foundations. With preliminary progresses achieved by early forms of pre-trained models (\eg, BERT~\cite{bert} and T5~\cite{t5}), people moved on to take advantage of LLMs for dense retrieval. In this direction, a pioneer work is made by OpenAI, where a series of GPT models were fine-tuned towards text and code representation~\cite{OpenAI-TextEmb}. It is the first time where decoding-only transformers were effectively applied for dense retrieval. Importantly, it empirically validates that the retrieval performance can be consistently improved from the increased model size and embedding dimension. Besides, the LLM-based retrievers also exhibit superior generalizability~\cite{Large-T5-Retriever,muennighoff2022sgpt}, as notable improvements can be achieved for not only the targeted scenario but also a variety of general tasks beyond the fine-tuned domain. 

Recently, the development of LLM-based dense retrievers have gotten dramatically promoted as powerful LLMs, \eg, LLaMA~\cite{llama}, Vicuna~\cite{vicuna}, Mistral~\cite{mistral}, Phi~\cite{phi}, and Gemma~\cite{gemma}, are made publicly available. Remarkably, RepLLaMA~\cite{rankllama}, the first fine-tuned embedder on top of open-source LLM (LLaMA-2-7B), brings forth major improvements on a variety of benchmarks, including MSMARCO passage/doc retrieval~\cite{MSMARCO} and BEIR~\cite{BEIR}. Despite extra computation costs due to the expanded model scale, the first-stage retrieval accuracy with RepLLaMA alone already surpasses the multi-stage retrieval accuracy resulted from conventional methods, indicating the its potential value for real-world application. After that, people make exploration of other alternatives for dense retrieval, where additional improvements are continually achieved with adoption of more advanced LLMs~\cite{GENQD,SFRAIResearch2024,LinqAIResearch2024}. To date, LLM-based embedders have dominated all major text retrieval benchmarks, \eg, currently, the leading methods on MTEB~\cite{muennighoff2022mteb} are all back-ended by LLMs. 

In addition to the above methods which directly fine-tune LLMs, there are also parallel works on adapting generic LLMs as better foundations for dense retrieval. For example, Llama2Vec \cite{li2023making} performs post pre-training of LLaMA-2 with two new pretext tasks: EBAE (embedding based auto-encoding) and EBAR (embedding based auto-regression). With moderate scale of training on unlabeled corpus, it results in substantial improvements of retrieval performance over the basic Llama-2 model. Besides, NV-Embed modifies LLM's architecture by introducing latent attention layer and bidirectional attention~\cite{lee2024nv}. Both modifications contribute to the improved performance on MTEB benchmark. Despite the above preliminary progresses, there are still many open challenges about LLM-based embedders, such as efficiency and adaptability, which need to be addressed in the future.

\mypara{Introducing New Capacities.}
Compared to conventional methods that use small-scale pre-trained models, LLM-based embedders introduce new capabilities that enhance the usability and accuracy of dense retrieval. One notable example is their ability to follow instructions, allowing LLM-based embedders to be trained for various semantic matching tasks based on user demands. For instance, an LLM-based embedder can perform document retrieval when prompted with ``\textit{retrieve relevant docs for the input question}'', and can be adapted for duplicate question retrieval with the prompt ``\textit{retrieve questions with the same meaning as input}''. Although BERT-based retrievers are also fine-tuned to follow instructions~\cite{su2022one,Task-aware-Instruction}, they do not support unseen instructions as effectively as LLM-based embedders~\cite{LLM4DR}. 
ChatRetriever~\cite{ChatRetriever} further leverages dialog-based instruction tuning to build LLM-based conversational embedders, enhancing their conversational retrieval capabilities.
In addition to instructions, the LLM-based embedders can also be adapted through in-context learning, where the retrieval function can be updated by demonstration examples of user's interested tasks~\cite{jiang2023scaling}. Another advantage of LLM-based embedders is their length-generalizable capacity, which allows them to effectively handle much longer texts than those in their training examples~\cite{LLM4DR}. This makes it possible to manage retrieval applications across various text lengths while maintaining a feasible training cost.


\subsubsection{Generative Retriever}
\label{gen_ir}
Traditional IR systems typically follow the ``index-retrieval-rank'' paradigm to locate relevant documents based on user queries. Though, this approach has proven its effective in practice, it usually consist of three separate modules: the index module, the retrieval module, and the reranking module. Optimizing these modules collectively can be challenging, potentially resulting in sub-optimal retrieval outcomes. Additionally, this paradigm demands additional storage space for pre-built indexes, further burdening storage resources.
Recently, generative model-based retrieval methods~\cite{model-based-ir-1, model-based-ir-2, model-based-ir-3} have emerged to address these challenges. These methods move away from the traditional ``index-retrieval-rank'' paradigm and instead use a unified model to directly generate document identifiers (\ie, DocIDs) relevant to the queries. In these generative retrieval methods, the knowledge of the document corpus is stored in the model parameters, eliminating the need for additional storage space for a separate index. Existing works have investigated the approaches of generating document identifiers through fine-tuning and prompting of LLMs~\cite{DSI,LLMurl} 

\mypara{Fine-tuning LLMs.} Given the vast amount of world knowledge contained in LLMs, it is intuitive to leverage them to build generative retrievers. DSI~\cite{DSI} is a typical method that fine-tunes the pre-trained T5 models on retrieval datasets. The approach involves encoding queries and decoding document identifiers directly to perform retrieval. They explore multiple techniques for generating document identifiers and find that constructing semantically structured identifiers yields optimal results. In this strategy, DSI applies hierarchical clustering to group documents according to their semantic embeddings and assigns a semantic DocID to each document based on its hierarchical group. To ensure the output DocIDs are valid and do represent actual documents in the corpus, DSI constructs a trie using all DocIDs and utilizes a constraint beam search during the decoding process. 
Furthermore, this approach observes that the scaling law, which suggests that larger LMs lead to improved performance, is also applied to generative retrievers. 
Though various generative retrievers have been proposed~\cite{DSI,NCI,ARGDI}, most of them mainly focus on fine-tuning size-limited LMs on small-size document corpus (usually a subset of MSMARCO~\cite{MSMARCO}). To analyze how the model size and document-corpus size impact the effectiveness of generative retrievers, \citet{GD-analyze} conducted a comprehensive analysis by scaling up corpus size from 100k to 8.8M and scaling model size up to 11B (T5-XXL). The primary findings are three-fold: (1) It is still challenging for generative retrievers to cover large-scale document corpus. (2) More model parameters often bring better performance. (3) Introducing synthetic queries generated from documents to expand training samples could significantly enhance the retrieval performance. 

CorpusLM~\cite{CorpusLM} further explores combining the retrieval and answering tasks together based on LLMs, making the two mutually reinforcing. Researchers devise various training tasks, \eg, DocID list generation, closed-book answers generation, and RAG generation, to sufficiently leverage and enhance the world knowledge of LLMs, improving the performance of these generation tasks. 

\mypara{Prompting LLMs.} In addition to fine-tuning LLMs for retrieval, it has been found that LLMs (\eg, GPT-series models) can directly generate relevant web URLs for user queries with a few in-context demonstrations~\cite{LLMurl}. This unique capability of LLMs is believed to arise from their training exposure to various HTML resources. As a result, LLMs can naturally serve as generative retrievers that directly generate document identifiers to retrieve relevant documents for input queries. To achieve this, an LLM-URL~\cite{LLMurl} model is proposed. It utilizes the GPT-3 \emph{text-davinci-003} model to yield candidate URLs. Furthermore, it designs regular expressions to extract valid URLs from these candidates to locate the retrieved documents.

To provide a comprehensive understanding of this topic, Table~\ref{tab:retrieve_model_comparison} summarizes the common and unique characteristics of the LLM-based retrievers discussed above.

%

\begin{table}[t]
    \centering
    \caption{The comparison of retrievers that leverage LLMs as the foundation. ``KD'' is short for ``Knowledge Distillation''.}
    \setlength{\tabcolsep}{1.5mm}{
    \begin{tabular}{lccccc}
    \toprule
    \textbf{Methods} & \textbf{Backbone} & \textbf{Architecture} & \textbf{LLM's tuning} \\
    \midrule
    cpt-text~\cite{OpenAI-TextEmb} & GPT-series & Dense & \makecell{Pre-training \&\\ Fine-tuning} \\
    GTR~\cite{Large-T5-Retriever} & T5 & Dense &  \makecell{Pre-training \& \\Fine-tuning} \\
    RepLLaMA~\cite{rankllama} & LLAMA & Dense & Fine-tuning \\
    TART-full~\cite{Task-aware-Instruction} & \makecell{T0 \& \\ Flan-T5} & Dense & \makecell{Fine-tuning \&\\ Prompting}\\
    TART-dual~\cite{Task-aware-Instruction} &  Contriever& Dense  & \makecell{KD \& \\ Prompting}\\
    DSI~\cite{DSI} & T5 & Generative  & Fine-tuning \\
    LLM-URL~\cite{LLMurl} & GPT-3 & Generative  & Prompting \\
    CorpusLM~\cite{CorpusLM} & \makecell{T5-base \&\\ Llama2-7B-Chat} & Generative &  Fine-tuning \\ 
    \bottomrule
    \end{tabular}
    }
    \label{tab:retrieve_model_comparison}
\end{table}

\subsection{Limitations}
Though some efforts have been made for LLM-augmented retrieval, there are still many areas that require more detailed investigation. For example, a critical requirement for retrievers is fast response, while the main problem of existing LLMs is the huge model parameters and overlong inference time. Addressing this limitation of LLMs to ensure the response time of retrievers is a critical task. Moreover, even when employing LLMs to augment datasets (a context with lower inference time demands), the potential mismatch between LLM-generated texts and real user queries could impact retrieval effectiveness. Furthermore, as LLMs usually lack domain-specific knowledge, they need to be fine-tuned on task-specific datasets before applying them to downstream tasks. Therefore, developing efficient strategies to fine-tune these LLMs with numerous parameters emerges as a key concern.

%% file: S5-reranker.tex
\section{Reranker}
\label{sec:rank}
Reranker, as the second-pass document filter in IR, aims to rerank a document list retrieved by the retriever (\eg, BM25) based on the query-document relevance. According to the usage of LLMs, existing LLM-based reranking methods can be divided into four paradigms: utilizing LLMs as supervised rerankers, utilizing LLMs as unsupervised rerankers, utilizing LLMs for training data augmentation and reasoning-intensive rerankers. These paradigms are summarized in Table~\ref{tab:reranker} and will be introduced in the following sections. Recall that we will use the term \textit{document} to refer to the text retrieved in general IR scenarios, including instances such as passages (\eg, passages in MS MARCO passage ranking dataset~\cite{MSMARCO}).

\begin{table}[t]
    \caption{Summary of existing LLM-based re-ranking methods. ``Enc'' and ``Dec'' denote encoder and decoder, respectively.}
    \label{tab:reranker}
    \setlength{\tabcolsep}{1.3mm}{
    \begin{tabular}{p{.2\linewidth} c p{.57\linewidth}}
    \toprule
    \textbf{Paradigm} & \textbf{Type} & \textbf{Method} \\ \midrule
    Supervised Rerankers & Enc-only & monoBERT~\cite{monobert} \\
     & Enc-dec & monoT5~\cite{monot5}, \citet{JuYW21}, DuoT5~\cite{Expando-Mono-Duo}, RankT5~\cite{RankT5}, ListT5~\cite{listT5} \\
     & Dec-only & RankLLaMA~\cite{rankllama}, TSARankLLM~\cite{TSARankLLM}, Q-PEFT~\cite{Q-PEFT}, \citet{Rank-without-GPT}, PE-Rank~\cite{PE-Rank} \\ \midrule
    Unsupervised Rerankers & Pointwise & \citet{relevance_generation}, \citet{beyond_yes_no}, \citet{Diverse_Criteria}, \citet{query_generation}, \citet{opensource}, \citet{Prompt_Variations}, Co-Prompt~\cite{Co-Prompt}, DemoRank~\cite{DemoRank}, PARADE~\cite{parade} \\
     & Listwise & RankGPT~\cite{sun2023chatgpt}, \citet{fullrank}, CoRanking~\cite{coranking}, \citet{ma2023zero}, \citet{self_consistency_llm}, TourRank~\cite{TourRank}, \citet{Top-Down}, APEER~\cite{apeer} \\
     & Pairwise & PRP~\cite{qin-etal-2024-large}, \citet{setwise}, PRP-Graph~\cite{PRP-Graph}, \citet{Post-Processing} \\ \midrule
    Data Augmentation & -- & ExaRanker~\cite{ExaRanker}, ExaRanker-Open~\cite{ExaRanker-Open}, InPars-Light~\cite{InPars-Light}, \citet{askari2023generating}, \citet{RL-driven}, RankVicuna~\cite{rankvicuna}, RankZephyr~\cite{rankzephyr}, \citet{sun2023instruction} \\ \midrule
    Reasoning-intensive Rerankers & -- & ReasonRank~\cite{reasonrank}, Rank1~\cite{rank1}, Rank-K~\cite{rank-k}, Rearank~\cite{rearank}, Rank-R1~\cite{rank-r1}, TFRank~\cite{tfrank} \\ \bottomrule
    \end{tabular}}
\end{table}

\subsection{Utilizing LLMs as Supervised Rerankers} \label{sec:supervised}
Supervised fine-tuning is an important step in applying pre-trained LLMs to a reranking task. Due to the lack of awareness of ranking during pre-training, LLMs cannot appropriately measure the query-document relevance and fully understand the reranking tasks. By fine-tuning LLMs on task-specific ranking datasets, such as the MS MARCO passage ranking dataset~\cite{MSMARCO}, which includes signals of both relevance and irrelevance, LLMs can adjust their parameters to yield better performance in the reranking tasks. Based on the backbone model structure, we can categorize existing supervised rerankers as: (1) encoder-only, (2) encoder-decoder, and (3) decoder-only.

\subsubsection{Encoder-only}
The encoder-based rerankers represent a significant turning point in applying LLMs to document reranking tasks. They demonstrate how some pre-trained language models~(\eg, BERT~\cite{bert}) can be fine-tuned to deliver highly accurate relevance predictions. A representative approach is monoBERT~\cite{monobert}, which transforms a query-document pair into a sequence ``[CLS] {\textit{query}} [SEP] {\textit{document}} [SEP]'' as the model input and calculates the relevance score by feeding the ``[CLS]'' representation into a linear layer. The reranking model is optimized based on the cross-entropy loss.

\subsubsection{Encoder-Decoder}
In this field, existing studies mainly formulate document reranking as a generation task and optimize an encoder-decoder-based reranking model~\cite{monot5, JuYW21, Expando-Mono-Duo, RankT5}. Specifically, given the query and the document, reranking models are usually fine-tuned to generate a single token, such as ``true'' or ``false''. During inference, the query-document relevance score is determined based on the logit of the generated token. For example, a T5 model can be fine-tuned to generate classification tokens for relevant or irrelevant query-document pairs~\cite{monot5}. At inference time, a softmax function is applied to the logits of ``true'' and ``false'' tokens, and the relevance score is calculated as the probability of the ``true'' token. The following method~\cite{JuYW21} involves a multi-view learning approach based on the T5 model. This approach simultaneously considers two tasks: generating classification tokens for a given query-document pair and generating the corresponding query conditioned on the provided document. DuoT5~\cite{Expando-Mono-Duo} considers a triple $(q, d_i, d_j)$ as the input of the T5 model and is fine-tuned to generate token ``true'' if document $d_i$ is more relevant to query $q_i$ than document $d_j$, and ``false'' otherwise. During inference, for each document $d_i$, it enumerates all other documents $d_j$ and uses global aggregation functions to generate the relevance score $s_i$ for document $d_i$ (\eg, $s_i = \sum_j p_{i,j}$, where $p_{i,j}$ represents the probability of generating ``true'' when taking $(q, d_i, d_j)$ as the model input). 

Beyond the aforementioned studies, several studies have also explored different training losses and model architectures for reranker training. For example, RankT5~\cite{RankT5} is proposed to directly yield a numerical relevance score for each query-document pair during training and optimize the ranking performance with ``pairwise'' or ``listwise'' ranking losses. It differs significantly from the previous studies that optimize rerankers by generating text tokens and using a generation loss, and the use of ranking loss (\eg, RankNet~\cite{RankNet}) is more reasonable and aligns better with the inherent nature of the ranking task. Besides, ~\citet{listT5} propose ListT5, a listwise reranker based on Fusion-in-decoder architecture. It jointly takes multiple documents as input and directly generates a reranked document list during training and inference.


\subsubsection{Decoder-only}\label{sec:rerank-decoder}
Recently, there have been some attempts~\cite{rankllama, TSARankLLM, Q-PEFT, Rank-without-GPT, PE-Rank} to rerank documents by fine-tuning decoder-only models (such as LLaMA). For example, RankLLaMA~\cite{rankllama} proposes formatting the query-document pair into a prompt ``query: \{\emph{query}\} document: \{\emph{document}\} [EOS]'' and utilizes the last token representation for relevance calculation. Besides, TSARankLLM~\cite{TSARankLLM} has been proposed to bridge the gap between LLMs' conventional training objectives and the specific needs of document reranking through two-stage training. The first stage involves continuously pretraining LLMs using a large number of relevant text pairs collected from web resources, helping the LLMs to naturally generate queries relevant to the input document. The second stage focuses on improving the model's text ranking performance using high-quality supervised data and well-designed loss functions. ~\citet{Q-PEFT} propose a query-dependent parameter efficient fine-tuning (Q-PEFT) approach for ranking, which helps the LLM generate true queries based on given documents. Different from these pointwise rerankers~\cite{rankllama, TSARankLLM, Q-PEFT}, ~\citet{Rank-without-GPT} and ~\citet{PE-Rank} proposes to train a listwise reranker that directly outputs a reranked document list. Specifically, ~\citet{Rank-without-GPT} first demonstrate that existing pointwise datasets (such as MS MARCO~\cite{MSMARCO}), which only contain binary query-document labels, are insufficient for training efficient listwise rerankers. Then, they propose to use the ranking results of existing ranking systems (such as Cohere rerank API) as gold rankings to train a listwise reranker based on Code-LLaMA-Instruct. ~\citet{PE-Rank} propose PE-Rank, which compresses each document in the list into a single embedding and then inputs these document embeddings into reranker, which significantly reduces the input length and improves the efficiency of reranker.

\begin{figure*}[t]
    \centering
    \includegraphics[width=.8\linewidth]{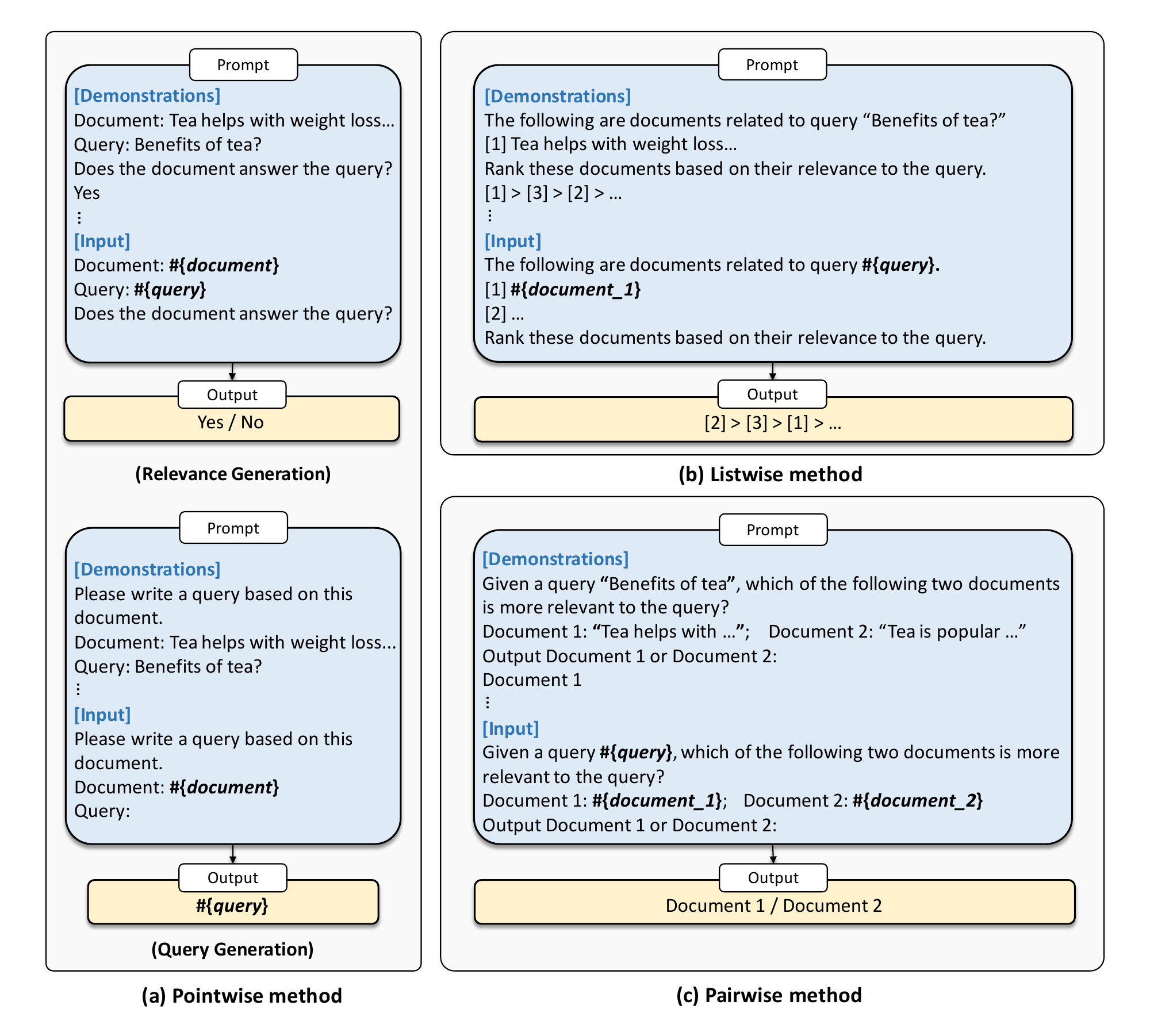}
    \caption{Three types of unsupervised reranking methods: (a) pointwise methods that consist of relevance generation (upper) and query generation (lower), (b) listwise methods, and (c) pairwise methods. The ``Demonstrations'' represents the few-shot demonstrations whose format are same as the current input.}
    \label{fig:reranking}
\end{figure*}

\subsection{Utilizing LLMs as Unsupervised Rerankers}
As the size of LLMs scales up (\eg, exceeding 10 billion parameters), it becomes increasingly difficult to fine-tune the reranking model. Addressing this challenge, recent efforts have attempted to prompt LLMs to directly enhance document reranking in an unsupervised way. In general, these prompting strategies can be divided into three categories: pointwise, listwise, and pairwise methods. A comprehensive exploration of these strategies follows in the subsequent sections.

\subsubsection{Pointwise methods}
The pointwise methods measure the relevance between a query and a single document, and can be categorized into two types: relevance generation~\cite{DemoRank, relevance_generation, beyond_yes_no, Diverse_Criteria} and query generation~\cite{query_generation, opensource, Co-Prompt}. 

The upper part in Figure~\ref{fig:reranking} (a) shows an example of relevance generation based on a given prompt, where LLMs output a binary label (``Yes'' or ``No'') based on whether the document is relevant to the query. Following~\cite{monot5}, the query-document relevance score $f(q, d)$ can be calculated based on the log-likelihood of token ``Yes'' and ``No'' with a softmax function:
\begin{equation}
    f(q, d) = \frac{\text{exp}(S_{Y})}{\text{exp}(S_{Y}) + \text{exp}(S_{N})},
\end{equation}
where $S_{Y}$ and $S_{N}$ represent the LLM’s log-likelihood scores of ``Yes'' and ``No'' respectively. In addition to binary labels, \citet{beyond_yes_no} propose to incorporate fine-grained relevance labels (\eg, ``highly relevant'', ``somewhat relevant'' and ``not relevant'') into the prompt, which helps LLMs more effectively differentiate among documents with varying levels of relevance to a query. ~\citet{Diverse_Criteria} discuss the issues of inconsistent and biased relevance assessments of existing pointwise rerankers and introduce MCRanker that generates relevance scores based on a series of criteria from multiple perspectives.

As for the query generation shown in the lower part of Figure~\ref{fig:reranking} (a), the query-document relevance score is determined by the average log-likelihood of generating the actual query tokens based on the document:
\begin{equation}
    \text{score} = \frac{1}{|q|} \sum_i \log p(q_i | q_{<i}, d, \mathcal{P}),
\end{equation}
where $|q|$ denotes the token number of query $q$, $d$ denotes the document, and $\mathcal{P}$ represents the provided prompt. The documents are then reranked based on their relevance scores. It has been proven that some LLMs (such as T0) yield significant performance in zero-shot document reranking based on the query generation method~\cite{query_generation}. Recently, research~\cite{opensource} has also shown that the LLMs that are pre-trained without any supervised instruction fine-tuning (such as LLaMA) also yield robust zero-shot ranking ability. 

Although effective, these methods primarily rely on a handcrafted prompt (\eg, ``Please write a query based on this document''), which may not be optimal. Previous study~\cite{Prompt_Variations} has shown that prompt has a significant impact on the performance of LLM reranker. Thus, how to design appropriate prompts for ranking task is an important problem. Along this line, a discrete prompt optimization method Co-Prompt~\cite{Co-Prompt} is proposed for better prompt generation in reranking tasks. Besides, PaRaDe~\cite{parade} proposes a difficulty-based method to select the most difficult $k$ in-context demonstrations to include in the prompt, proving improvements compared with zero-shot performance. Nevertheless, the experiments in the paper indicate that such difficulty-based selection does not even show a significant advantage compared to random selection, showing that the demonstration selection in ranking task is a very challenging problem. The main challenge lies in the complex nature of query-document relationship, which requires effectively combining multiple demonstrations to help the LLM understand such relationship. Aiming to select more effective demonstrations for ranking task, ~\citet{DemoRank} propose DemoRank, an effective demonstration selection framework. The core component of DemoRank is a dependency-aware demonstration reranker, which reranks a list of demonstrations (usually obtained by a demonstration retriever) so that the combination of top-ranked demonstrations can yield better performance. An efficient method is proposed to construct the training samples for such demonstration reranker and a novel list-pairwise training loss is designed for optimization.

\begin{table*}[t]
\centering
\small
\caption{The comparison between different LLM-based reranking methods. $N$ denotes the number of documents to rerank. The ``complexity'', ``logits'', and ``batch'' represent the computational complexity, whether accesses LLM's logits, and whether allows batch inference respectively. $k$ is the constant in sliding windows strategy. As for the performance, we use NDCG@10 as a metric, and the results are calculated by reranking the top-100 documents retrieved by BM25 on TREC-DL2019 and TREC-DL2020. The best model is in bold while the second-best is underlined. The results come from previous study~\cite{qin-etal-2024-large}. *Since the parameters of ChatGPT have not been released, its model parameters are based on public estimates~\cite{baktash2023gpt}.}
\label{tab:reranker_comparison}
\setlength{\tabcolsep}{1.4mm}{
\begin{tabular}{llllccccc}
\toprule
\multirow{2}{*}{} & \multirow{2}{*}{\textbf{Method}} & \multirow{2}{*}{\textbf{LLM}} & \multirow{2}{*}{\textbf{Size}} & \multicolumn{3}{c}{\textbf{Properties}} & \multicolumn{2}{c}{\textbf{Performance}} \\ \cmidrule(lr){5-7} \cmidrule(lr){8-9}
 &  &  &  & \textbf{Complexity} & \textbf{Logits} & \textbf{Batching} & \textbf{TREC-DL19} & \textbf{-DL20} \\ \midrule
Initial Retriever & BM25 & - & - & - & - & - & 50.58 & 47.96 \\ \midrule
\multirow{3}{*}{Supervised} & monoBERT~\cite{monobert} & BERT & 340M & - & $\checkmark$ & $\checkmark$ & 70.50 & 67.28 \\
 & monoT5~\cite{monot5} & T5 & 220M & - & $\checkmark$ & $\checkmark$ & 71.48 & 66.99 \\
 & RankT5~\cite{RankT5} & T5 & 3B & - & $\checkmark$ & $\checkmark$ & 71.22 & 69.49 \\ \midrule
\multirow{2}{*}{\makecell[l]{Unsupervised \\ Pointwise}} & Query Generation~\cite{query_generation} & FLAN-UL2 & 20B & $O(N)$ & $\checkmark$ & $\checkmark$ & 58.95 & 60.02 \\
 & Relevance Generation~\cite{relevance_generation} & FLAN-UL2 & 20B & $O(N)$ & $\checkmark$ & $\checkmark$ & 64.61 & 65.39 \\
\multirow{2}{*}{\makecell[l]{Unsupervised \\ Listwise}} & RankGPT$_{3.5}$~\cite{sun2023chatgpt} & ChatGPT & 154B* & $O(k*N)$ &  &  & 65.80 & 62.91 \\
 & RankGPT$_4$~\cite{sun2023chatgpt} & GPT-4 & 1T* & $O(k*N)$ &  &  & \textbf{75.59} & \underline{70.56} \\
\multirow{2}{*}{\makecell[l]{Unsupervised \\ Pairwise}} & PRP-Allpair~\cite{qin-etal-2024-large} & FLAN-UL2 & 20B & $O(N^2)$ & $\checkmark$ & $\checkmark$ & \underline{72.42} & \textbf{70.68} \\
 & PRP-Heapsort~\cite{qin-etal-2024-large} & FLAN-UL2 & 20B & $O(N*logN)$ & $\checkmark$ &  & 71.88 & 69.43 \\ \bottomrule
\end{tabular}}
\end{table*}

\subsubsection{Listwise Methods}
Listwise methods~\cite{sun2023chatgpt, ma2023zero} aim to directly rank a list of documents (see Figure~\ref{fig:reranking} (b)). These methods insert the query and a document list into the prompt and instruct the LLMs to output the reranked document identifiers. Due to the limited input length of LLMs, it is not feasible to insert all candidate documents into the prompt. To alleviate this issue, these methods employ a sliding window strategy to rerank a subset of candidate documents each time. This strategy involves ranking from back to front using a sliding window, re-ranking only the documents within the window at a time. 

Although listwise methods have yielded promising performance, they still suffer from some weaknesses: (1) The performance of listwise methods is highly sensitive to the document order in the prompt. When the document order is randomly shuffled, listwise methods perform even worse than BM25~\cite{sun2023chatgpt}, revealing positional bias issues in the listwise ranking of LLMs. (2) The use of sliding windows limits the number of documents that can be ranked each time, and the dependency between adjacent windows prevents parallelization of LLM inference, thereby reducing the efficiency of reranking. Recently, some studies have attempted to mitigate these issues. \citet{self_consistency_llm} introduce a permutation self-consistency method, which involves shuffling the list in the prompt and aggregating the generated results to achieve a more accurate and a positionally unbiased ranking. ~\citet{TourRank} introduce a tournament mechanism into listwise ranking and propose TourRank, which parallelizes the reranking process through intelligent grouping and use a tournament-like points system to reduce the impact of the initial document order. ~\citet{Top-Down} propose a parallelizable partitioning algorithm for listwise ranking, which also aims at mitigating efficiency issues. ~\citet{FIRST} propose a novel listwise reranking approach which leverages the output logits of the first generated identifier to accelerating reranking process. To optimize the listwise reranking prompt, ~\citet{apeer} propose a novel automatic prompt
engineering algorithm APEER, which generates prompts through feedback and preference optimization. ~\citet{fullrank} comprehensively discuss the benefits of long-context LLMs for listwise ranking and introduce a novel full reranker which performs better than the sliding window reranker while also being more efficient and having lower API cost. ~\citet{coranking} propose a collaborative ranking framework CoRanking which combines small and large listwise rerankers for more efficient and effective passage ranking. They also design a novel passage order adjuster to mitigate the sensitivity of listwise reranker to the input document order.


\subsubsection{Pairwise Methods}
In pairwise methods, LLMs are given a prompt that consists of a query and a document pair (see Figure~\ref{fig:reranking} (c)). Then, they are instructed to generate the identifier of the document with higher relevance. To rerank all candidate documents, aggregation methods like AllPairs are used~\cite{qin-etal-2024-large}. AllPairs first generates all possible document pairs, yields discrete judgments for each pair (\eg, Document 1 or Document 2), and aggregates a final relevance score for each document. Efficient sorting algorithms, such as heap sort and bubble sort, are employed to speed up the ranking process. These sorting algorithms utilize efficient data structures to compare document pairs selectively and elevate the most relevant documents to the top of the ranking list, which is particularly useful in top-$k$ ranking. Experimental results show the state-of-the-art performance on the standard benchmarks using moderate-size LLMs (\eg, Flan-UL2 with 20B parameters), which are much smaller than those typically employed in listwise methods (\eg, GPT3.5). Building on the pairwise prompting approach, several ranking method variants have been proposed. ~\citet{PRP-Graph} introduce an innovative scoring unit that leverages the generation probability of judgments instead of discrete judgments, and further design a graph-based aggregation approach to obtain a final relevance score for each document. ~\citet{SinhababuPGSM24} and  propose to utilize few-shot in-context demonstrations to improve the performance of pairwise ranking. ~\citet{Post-Processing} utilize the pairwise comparison as a post-processing step to adjust the relevance scores generated by the pointwise LLM reranker.

Although effective, pairwise methods still suffer from high time complexity. To alleviate the efficiency problem, a setwise approach~\cite{setwise} has been proposed to compare a set of documents at a time and select the most relevant one from them. This approach allows the sorting algorithms (such as heap sort) to compare more than two documents at each step, thereby reducing the total number of comparisons and speeding up the sorting process.

\subsubsection{Comparison and Discussion}
We compare different unsupervised methods from various aspects to better illustrate the strengths and weaknesses of each method, which is summarized in Table~\ref{tab:reranker_comparison}. The representative methods~\cite{query_generation, relevance_generation, sun2023chatgpt, qin-etal-2024-large} in pointwise, listwise, and pairwise ranking, and some supervised methods~\cite{monobert, monot5, RankT5} mentioned in Section~\ref{sec:supervised} are selected for performance comparison. 

The pointwise methods (query generation and relevance generation) judge the relevance of each query-document pair independently, thus offering lower time complexity and enabling batch inference. However, compared to other methods, it does not have an advantage in terms of performance. The listwise method yields significant performance especially when calling GPT-4, but suffers from expensive API cost and non-reproducibility~\cite{rankvicuna}. Compared with the listwise method, the pairwise method shows competitive results based on a much smaller model FLAN-UL2 (20B). Stemming from the necessity to compare an extensive number of document pairs, its primary drawback is low efficiency.

\subsection{Utilizing LLMs for Training Data Augmentation}
Furthermore, in the realm of reranking, researchers have explored the integration of LLMs for training data augmentation~\cite{ExaRanker, InPars-Light, askari2023generating, rankvicuna, rankzephyr, sun2023instruction}. For example, ExaRanker~\cite{ExaRanker} and ExaRanker-Open~\cite{ExaRanker-Open} generate explanations for query-passage pairs using GPT-3.5 and open-source LLMs respectively, and subsequently trains a seq2seq ranking model to generate relevance labels along with corresponding explanations. InPars-Light~\cite{InPars-Light} is proposed as a cost-effective method to synthesize queries for documents by prompting LLMs. ~\citet{askari2023generating} proposes to generate synthetic documents based on LLMs in response to user queries. Furthermore, ~\citet{RL-driven} propose to utilize reinforcement learning to improve the quality of synthetic documents generated by LLMs.

Recently, many studies~\cite{rankvicuna, rankzephyr, sun2023instruction} have also attempted to distill the document reranking capability of LLMs into a specialized model. RankVicuna~\cite{rankvicuna} proposes to use the ranking list of $\text{RankGPT}_{3.5}$~\cite{sun2023chatgpt} as the gold list to train a 7B parameter Vicuna model. RankZephyr~\cite{rankzephyr} introduces a two-stage training strategy for distillation: initially applying the RankVicuna recipe to train $\text{Zephyr}{\gamma}$ in the first stage, and then further finetuning it in the second stage with the ranking results from $\text{RankGPT}_{4}$. These two studies not only demonstrate competitive results but also alleviate the issue of ranking results non-reproducibility of black-box LLMs. Besides, researchers~\cite{sun2023instruction} have also tried to distill the ranking ability of a pairwise ranker, which is computationally demanding, into a simpler but more efficient pointwise ranker.

\subsection{Reasoning-intensive Rerankers}
\label{sec:reasoning-rerankers}
Recent breakthroughs in Large Reasoning Models (LRMs) like DeepSeek-R1~\cite{r1} have demonstrated exceptional capabilities across many NLP tasks. These models significantly improve the answer accuracy in many complex NLP tasks (\eg, math and coding) through explicit step-by-step reasoning chains during inference. This capability holds particular promise for document reranking, where precise understanding of query intent and cross-document comparison are critical for relevance assessment. 

Motivated by these advancements, emerging research has explored injecting reasoning ability into document rerankers. For example, ~\citet{rank1} and ~\citet{rank-k} propose to apply DeepSeek-R1 as a teacher model to distill its reasoning process into smaller rerankers. ~\citet{rearank} and ~\citet{rank-r1} propose to use reinforcement learning algorithm to optimize reranker based on rule-based reward. TFRank~\cite{tfrank} introduces a ``think-free'' pointwise ranker that leverages reasoning during training while eliminating intermediate reasoning steps at inference, significantly improving the reasoning efficiency. While effective, these rerankers are primarily trained on traditional web search data MSMARCO, making them difficult to generalize to many complex and reasoning-intensive ranking benchmarks~\cite{bright}. To address the scarcity of reasoning-intensive training data, \citet{reasonrank} propose an automated data synthesis framework and generate 13K high-quality reasoning-intensive training data covering diverse search scenarios. They further propose a two-stage ``SFT+RL'' training framework, to empower LLM with strong reasoning and ranking abilities. Their ReasonRank model has achieved state-of-the-art performance on many reasoning-intensive IR benchmarks such as BRIGHT~\cite{bright} and R2MED~\cite{r2med}.

\subsection{Limitations}
Although recent research on utilizing LLMs for document reranking has made significant progress, it still has some limitations. First, due to the reliance on API calls and a large number of parameters, the process of LLM ranking could be expensive and inefficient. Therefore, achieving a trade-off between the cost/efficiency and performance of LLMs is a topic worth discussing. Along this line, ~\citet{ecorank} propose a budget-aware ranking solution which maximizes the LLM's performance within a given budget. Notably, \citet{ChenGS25} introduce in-context re-ranking (ICR), an attention-based method that achieves superior efficiency by eliminating generative overhead through O(1) forward passes. Besides, ~\citet{MengAAAR24} systematically discuss the improvements in ranking efficiency and effectiveness brought by the rank list truncation technique. Second, while existing studies mainly focus on applying LLMs to open-domain datasets (such as MSMARCO~\cite{MSMARCO}) or relevance-based text ranking tasks, their adaptability to in-domain datasets~\cite{BEIR}, non-standard ranking datasets~\cite{WachsmuthSS18} and reasoning-intensive datasets~\cite{bright} remains an area that demands more comprehensive exploration.

%% file: S6-reader.tex
\section{Reader}
\begin{table*}[t]
    \centering
    \caption{The comparison of existing representative methods that have a passive reader module. REALM and RAG do not use LLMs, but their frameworks have been widely applied in many following approaches.}
    \setlength{\tabcolsep}{1.6mm}{
    \begin{tabular}{lcccc}
    \toprule
        \textbf{Methods} & \textbf{Backbone models} & \textbf{Where to incorporate retrieval} & \textbf{When to retrieve} & \textbf{How to use LLMs} \\
    \midrule
        REALM~\cite{REALM} & BERT & Input layer & In the beginning & Fine-tuning \\ 
        RAG~\cite{RAG} & BART & Input layer & In the beginning & Fine-tuning \\
        REPLUG~\cite{REPLUG} & GPT & Input layer & In the beginning & Fine-tuning\\
        Atlas~\cite{Atlas} & T5 & Input layer & In the beginning & Fine-tuning \\
        \citet{RAG1} & Gopher & Input layer & In the beginning & Prompting \\
        \citet{RAG2} & GPT & Input layer & In the beginning & Prompting \\
        Chain-of-Note~\cite{chain-of-note} & LLaMA & Input layer & In the beginning & Fine-tuning \\
        RALM~\cite{RALM} & LLaMA \& OPT \& GPT & Input layer & During generation (every $n$ tokens) & Prompting \\
        RETRO~\cite{RETRO} & Transformer & Attention layer & During generation (every $n$ tokens) & Training from scratch \\
        ITERGEN~\cite{itergen2} & GPT & Input layer & During generation (every answer) & Prompting\\
        IRCoT~\cite{IRCOT} & Flan-T5 \& GPT & Input layer & During generation (every sentence) & Prompting \\
        FLARE~\cite{ARG} & GPT & Input layer & During generation (aperiodic) & Prompting \\
        Self-RAG~\cite{self-RAG} & LLaMA & Input layer & During generation (aperiodic) & Fine-tuning \\
    \bottomrule
    \end{tabular}
    }
    \label{tab:reader}
\end{table*}

\label{sec:reader}
With the impressive capabilities of LLMs in understanding, extracting, and processing textual data, researchers explore expanding the scope of IR systems beyond content ranking to answer generation. In this evolution, a reader module has been introduced to generate answers based on the document corpus in IR systems. By integrating a reader module, IR systems can directly present conclusive passages to users. Compared with providing a list of documents, users can simply comprehend the answering passages instead of analyzing the ranking list in this new paradigm. Furthermore, by repeatedly providing documents to LLMs based on their generating texts, the final generated answers can potentially be more accurate and information-rich than the original retrieved lists. 

A naive strategy for implementing this function is to heuristically provide LLMs with documents relevant to the user queries or the previously generated texts to support the following generation. However, this passive approach limits LLMs to merely collecting documents from IR systems without active engagement. An alternative solution is to train LLMs to interact proactively with search engines. For example, LLMs can formulate their own queries instead of relying solely on user queries or generated texts for references. According to the way LLMs utilize IR systems in the reader module, we can categorize them into \textit{passive readers} and \textit{active readers}. Each approach has its advantages and challenges for implementing LLM-powered answer generation in IR systems. Furthermore, since the documents provided by upstream IR systems are sometimes too long to directly feed as input for LLMs, some compression modules are proposed to extractively or abstractively compress the retrieved contexts for LLMs to understand and generate answers for queries. We will present these reader and compressor modules in the following parts and briefly introduce the existing analysis work on retrieval-augmented generation (RAG) strategy and their applications.

\subsection{Passive Reader}
To generate answers for users, a straightforward strategy is to supply the retrieved documents according to the queries or previously generated texts from IR systems as inputs to LLMs for creating passages~\cite{retallm, REALM, REPLUG, Atlas, RAG1, RAG2, RAG3, importancerag, IRCOT, RAG, RALM, RETRO, ARG, richrag, RoleRAG}. By this means, these approaches use the LLMs and IR systems separately, with LLMs functioning as passive recipients of documents from the IR systems. The strategies for utilizing LLMs within IR systems' reader modules can be categorized into the following three groups according to the frequency of retrieving documents for LLMs.

\subsubsection{Once-Retrieval Reader}
To obtain useful references for LLMs to generate responses for user queries, an intuitive way is to retrieve the top documents based on the queries themselves in the beginning. For example, REALM~\cite{REALM} adopts this strategy by directly attending the document contents to the original queries to predict the final answers based on masked language modeling. RAG~\cite{RAG} follows this strategy but applies the generative language modeling paradigm. However, these two approaches only use language models with limited parameters, such as BERT and BART. Recent approaches such as REPLUG~\cite{REPLUG} and Atlas~\cite{Atlas} have improved them by leveraging LLMs such as GPTs, T5s, and LLaMAs for response generation. To yield better answer generation performances, these models usually fine-tune LLMs on QA tasks. However, due to the limited computing resources, many methods~\cite{RAG1,RAG2, RAG3, longrag} choose to prompt LLMs for generation as they could use larger LMs in this way. Furthermore, to improve the quality of the generated answers, several approaches~\cite{ALCE, chain-of-note} also try to train or prompt the LLMs to generate contexts such as citations or notes in addition to the answers to force LLMs to understand and assess the relevance of retrieved passages to the user queries. ActiveRAG~\cite{activerag} and PG-RAG~\cite{PG-RAG} improve them by using knowledge construction during the answer generation process. Some approaches~\cite{importancerag, REAR} evaluate the importance of each retrieved reference using policy gradients to indicate which reference is more useful for generating. Specifically, ~\cite{REAR} utilize LLMs themselves to provide importance for different references which also supply additional training signals. Besides, researchers explore instruction tuning LLMs such LLaMAs to improve their abilities to generate conclusive passages relying on retrieved knowledge~\cite{SAIL,RADIT,instructrag}. During the training of LLM-based readers, some approaches~\cite{contras_reader} explore the strategy of contrastive learning by augmenting training data by removing and replacing retrieved passages to improve the generating performances. Additionally, SPRING~\cite{springrag} inserts several trainable tokens between the retrieved documents and issued questions for better optimization of the reader. R$^2$AG~\cite{R2AG} extracts features from retrieval models and attaches them to the reference contents to overcome the semantic gaps between LLMs and retrievers. ~\citet{irr_rag} also propose to generate noisy training data to help LLMs generate correct answers while irrelevant contents are included in the retrieved contexts. RAAT~\cite{noisy_rag} and ATM~\cite{ATM} further solve the noisy problem by introducing the adversarial training strategy.

\subsubsection{Periodic-Retrieval Reader}
However, while generating long conclusive answers, it is shown~\cite{RETRO, RALM} that only using the references retrieved by the original user intents as in once-retrieval readers may be inadequate. For example, when providing a passage about ``Barack Obama'', language models may need additional knowledge about his university, which may not be included in the results of simply searching the initial query. In conclusion, language models may need extra references to support the following generation during the generating process, where multiple retrieval processes may be required. 
To address this, solutions such as RETRO~\cite{RETRO} and RALM~\cite{RALM} have emerged, emphasizing the periodic collection of documents based on both the original queries and the concurrently generated texts (triggering a retrieval every $n$ generated tokens). In this manner, when generating the text about the university career of Barack Obama, the LLM can receive additional documents as supplementary materials. This need for additional references highlights the necessity for multiple retrieval iterations to ensure robustness in subsequent answer generation. Notably, RETRO~\cite{RETRO} introduces a novel approach incorporating cross-attention between the generating texts and the references within the Transformer attention calculation, as opposed to directly embedding references into the input texts of LLMs. Since it involves additional cross-attention modules in the Transformer's structure, RETRO trains this model from scratch. However, these two approaches mainly rely on the successive $n$ tokens to separate generation and retrieve documents, which may not be semantically continuous and may cause the collected references noisy and useless. To solve this problem, some approaches such as IRCoT~\cite{IRCOT} also explore retrieving documents for every generated sentence, which is a more complete semantic structure. Furthermore, researchers find that the whole generated passages can be considered as conclusive contexts for current queries and can be used to find more relevant knowledge to generate more thorough answers. Consequently, many recent approaches~\cite{itergen1, itergen2, itergen3, IM-RAG, GRG} have also tried to extend this periodic-retrieval paradigm to iteratively using the whole generated passages to retrieve references to re-generate the answers, until the iterations reach a pre-defined limitation. Particularly, these methods can be regarded as special periodic-retrieval readers that retrieve passages when every answer is (re)-generated. Since the LLMs can receive more comprehensive and relevant references with the iterations increase, these methods that combine RAG and generation-augmented retrieval strategies can generate more accurate answers but consume more computation costs.

\subsubsection{Aperiodic-Retrieval Reader}
In the above strategy, the retrieval systems supply documents to LLMs in a periodic manner. However, retrieving documents in a mandatory frequency may mismatch the retrieval timing and can be costly. Recently, FLARE~\cite{ARG} has addressed this problem by automatically determining the timing of retrieval according to the probability of generating texts. Since the probability can serve as an indicator of LLMs' confidence during text generation~\cite{DBLP:journals/corr/abs-2207-05221, DBLP:journals/tacl/JiangADN21}, a low probability for a generated term could suggest that LLMs require additional knowledge. Specifically, when the probability of a term falls below a predefined threshold, FLARE employs IR systems to retrieve references in accordance with the ongoing generated sentences, while removing these low-probability terms. FLARE adopts this strategy of prompting LLMs for answer generation solely based on the probabilities of generating terms, avoiding the need for fine-tuning while still maintaining effectiveness. Besides, self-RAG~\cite{self-RAG} tends to solve this problem by training LLMs such as LlaMA to generate specific tokens when they need additional knowledge to support following generations. Another critical model is introduced to judge whether the retrieved references are beneficial for generating.

We summarize representative passive reader approaches in Table~\ref{tab:reader}, considering various aspects such as the backbone language models, the insertion point for retrieved references, the timing of using retrieval models, and the tuning strategy employed for LLMs.

\subsection{Active Reader}
However, the passive reader-based approaches separate IR systems and generative language models. This signifies that LLMs can only submissively utilize references provided by IR systems and are unable to interactively engage with the IR systems in a manner akin to human interaction such as issuing queries to seek information.

To allow LLMs to actively use search engines, Self-Ask~\cite{Self-Ask}, DSP~\cite{DSP}, and PlanRAG~\cite{PlanRAG} try to employ few-shot prompts for LLMs, triggering them to search queries when they believe it is required. For example, in a scenario where the query is \textit{``When was the existing tallest wooden lattice tower built?''}, these prompted LLMs can decide to search a query \textit{``What is the existing tallest wooden lattice tower''} to gather necessary references as they find the query cannot be directly answered. Once acquired information about the tower, they can iteratively query IR systems for more details until they determine to generate the final answers instead of asking questions. To alleviate the problem of insufficient manually annotated data for fine-tuning, LPKG~\cite{LPKG} constructs high-quality active retrieval-augmented reasoning paths from existing knowledge graphs. Notably, these methods involve IR systems to construct a single reasoning chain for LLMs. MRC~\cite{MRC} further improves these methods by prompting LLMs to explore multiple reasoning chains and subsequently combining all generated answers using LLMs.

\subsection{Compressor}

Existing LLMs, especially open-sourced ones, such as LLaMA and Flan-T5, have limited input lengths (usually 4,096 or 8,192 tokens). However, the documents or web pages retrieved by upstream IR systems are usually long. Therefore, it is difficult to concatenate all the retrieved documents and feed them into LLMs to generate answers. Though some approaches manage to solve these problems by aggregating the answers supported by each reference as the final answers, this strategy neglects the potential relations between retrieved passages. A more straightforward way is to directly compress the retrieved documents into short input tokens or even dense vectors~\cite{LeanContext, RECOMP, FILCO, TCRA, PRCA, xRAG, bider}.

To compress the retrieved references, an intuitive idea is to extract the most useful $K$ sentences from the retrieved documents. LeanContext~\cite{LeanContext} applies this method and trains a small model by reinforcement learning (RL) to select the top $K$ similar sentences to the queries. The researchers also augment this strategy by using a free open-sourced text reduction method for the rest sentences as a supplement. Instead of using RL-based methods, RECOMP~\cite{RECOMP} directly uses the probability or the match ratio of the generated answers to the golden answers as signals to build training datasets and tune the compressor model. For example, the sentence corresponding to the highest generating probability is the positive one while others are negative ones. Furthermore, FILCO~\cite{FILCO} applies the ``hindsight'' methods, which directly align the prior distribution (the predicted importance probability distribution of sentences without knowing the gold answer) to the posterior distribution (the same distribution of sentences within knowing the gold answer) to tune language models to select sentences.

However, these extractive methods may lose potential intent among all references. Therefore, abstractive methods are proposed to summarize retrieved documents into short but concise summaries for downstream generation.  These methods~\cite{TCRA, RECOMP} usually distill the summarizing abilities of LLMs to small models. For example, TCRA~\cite{TCRA} leverages GPT-3.5-turbo to build abstractive compression datasets for MT5 model. Recently, xRAG~\cite{xRAG} proposes to use a freeze sentence encoder and tunes a projector to comprise retrieved passage into a dense vector.

\subsection{Analysis}
With the rapid development of the above reader approaches, many researchers have begun to analyze the characteristics of retrieval-augmented LLMs:

$\bullet$ \citet{lost_in_middle} find that the position of the relevant/golden reference has significant influences on the final generation performance. The performance is always better when the relevant reference is at the beginning or the end, which indicates the necessity of introducing a ranking module to order the retrieved knowledge. 

$\bullet$ \citet{LLMbond} observe that by applying retrieval augmentation generation strategy, LLMs can have a better awareness of their knowledge boundaries. 

$\bullet$ \citet{integration_str} analyze different strategies of integrating retrieval systems and LLMs such as concatenate (\ie, concatenating all references for answer generation) and post fusion (\ie, aggregating the answers corresponding to each reference). They also explore several ways of combining these two strategies.

$\bullet$ \citet{rag_tradeoff} demonstrate that there exists an attribution and fluency tradeoff for retrieval-augmented LLMs: with more received references, the attribution of generated answers increases while the fluency decreases.

$\bullet$ \citet{if_rag} argue that always retrieving references to support LLMs to generate answers hurts the question-answering performance. The reason is that LLMs themselves may have adequate knowledge while answering questions about popular entities and the retrieved noisy passages may interfere and bias the answering process. To overcome this challenge, they devise a simple strategy that only retrieves references while the popularity of entities in the query is quite low. By this means, the efficacy and efficiency of RAG both improve. \citet{ifrag2} pay attention to the same phenomenon and propose to paraphrase several perturbed questions for LLMs to answer according to their internal knowledge and perform a consistency check to decide whether to retrieve external information. ~\cite{ifrag3,ifrag4,ifrag5} also focus on this problem using triplets extracted from the knowledge graph and the confidence of LLMs. ~\cite{ifrag6, ifrag7} solve this problem by training LLMs or small language models to judge whether the questions are known by LLMs.

$\bullet$ \citet{knowledge-conflict} analyze the impacts of knowledge conflict among retrieved references and LLM's internal knowledge. and find that LLMs follow the majority rule while facing this phenomenon.

$\bullet$ \citet{rag_attack}, \citet{rag_attack2}, and~\citet{phantom} explore the attacking technique towards LLM-based retrieval augmented generation by poisoning retrieved passages. They find that even introducing some typos in the references may also affect the answer generation.

$\bullet$ \citet{domainrag} construct an in-domain reader evaluation dataset. They deeply analyze the effectiveness of the retrieval augmented generation paradigm under the long-tail and in-domain situations.

$\bullet$ \citet{reader_compare} compare the performances between readers based on base LLMs and ``instructed'' LLMs. Different from previous popular belief, They find base models outperform their corresponding instruction-tuned versions.

\subsection{Applications}
Recently, researchers~\cite{readerapp1, readerapp2, readerapp3, readerapp4, readerapp5, readerapp6, M-RAG} have applied the RAG strategy to areas such as clinical QA, medical QA, and financial QA to enhance LLMs with external knowledge and to develop domain-specific applications. For example, ATLANTIC~\cite{readerapp2} adapts Atlas to the scientific domain to derive a science QA system. Besides, some approaches~\cite{PRAG} also apply techniques in federated learning such as multi-party computation to perform personal RAG with privacy protection. 

Furthermore, to better facilitate the deployment of these RAG systems, some tools or frameworks are proposed~\cite{retallm, readertool1, readertool2}. For example, RETA-LLM~\cite{retallm} breaks down the whole complex generation task into several simple modules in the reader pipeline. These modules include a query rewriter module for refining query intents, a passage extraction module for aligning reference lengths with LLM limitations, and a fact verification module for confirming the absence of fabricated information in the generated answers. \citet{flashrag}~release the FlashRAG toolkit for the reproduction and development of RAG research, which includes 32 pre-processed benchmark datasets and 14 state-of-the-art algorithms.

\subsection{Limitations}
Several IR systems applying the RAG strategy, such as {New Bing} and {Langchain}, have already entered commercial use. However, there are also some challenges in this novel retrieval-augmented content generation system. These include challenges such as effective query reformulation, optimal retrieval frequency, correct document comprehension, accurate passage extraction, and effective content summarization. It is crucial to address these challenges to effectively realize the potential of LLMs in this paradigm.

%% file: S7-agent.tex
\section{Search Agent} \label{sec:agent}
The emergence of large reasoning models (LRMs) has ushered in a new era for IR systems, with a growing focus on developing LRM-based intelligent agents. This paradigm shift seeks to replicate human-like reasoning and retrieval processes, thereby augmenting the capacity of LLM-powered IR models to tackle complex, real-world problems. Leveraging their advanced natural language understanding, reasoning, and generation capabilities, these agents can autonomously search, interpret, and synthesize information from diverse sources.

Initial research in this domain focused on static pipeline-based architectures, where an information-seeking task is broken down into a series of modules, each with a pre-defined role~\cite{agent/static/lamda,agent/static/seeker,agent/static/webglm,agent/static/webagent,agent/static/gophercite,agent/static/knowwheretogo,CoSearchAgent}. While these systems demonstrate a foundational approach, their fixed workflows limit their ability to adapt to the dynamic and complex interactions inherent in real-world scenarios. This inflexibility constrains their overall performance and hinders their effectiveness in advanced reasoning and problem-solving.


Recently, the development of LRMs has enabled the development of a new class of autonomous search agents. These agents move beyond static pipelines by allowing the LLM to actively and dynamically explore the web. This is achieved by enabling the model to decide its next action based on real-time feedback from the environment or humans. This shift towards flexible, self-guided behavior makes these agents more adaptable and more closely aligned with human-like problem-solving.


In this section, we will comprehensively introduce the studies about search agents from the following four aspects: (1) architecture of search agents, (2) information seeking module, (3) optimization of search agents, and (4) benchmarks and resources.

\subsection{Architecture of Search Agent}\label{sec:sa_workflow}

The design of a search agent's architecture is a foundational step that establishes its core operational mechanism. Existing approaches can be broadly categorized into two main paradigms: \textit{single-agent frameworks} and \textit{multi-agent frameworks}. A single-agent framework utilizes a single LLM to handle all aspects of the task, including reasoning, interaction, and answer generation. In contrast, a multi-agent framework distributes these responsibilities among multiple LLMs, which act as collaborative agents to achieve the target.


\subsubsection{Single-Agent Frameworks}
In a single-agent framework, a single LLM with strong reasoning capabilities serves as the central decision-maker and task executor. This LLM dynamically determines the next action based on the current information state and environmental context, managing the entire reasoning and interaction process.
For example, Search-R1~\cite{Search-R1} and ReSearch~\cite{ReSearch} adopt a ReAct-style mechanism, enabling the LLM policy to automatically generate actions such as ``think'', ``search'', and ``answer''. This allows the agent to iteratively interact with search tools to resolve complex multi-hop questions. To optimize the performance of these single-agent systems, researchers have employed various reinforcement learning (RL) techniques. The GRPO algorithm~\cite{deepseek-math} has been commonly utilized to enhance performance. In a more refined approach, R1-Searcher~\cite{R1-Searcher} introduces a two-stage GRPO-based RL optimization. The first stage assigns rewards based on retrieval frequency and output format correctness, while the second stage rewards answer accuracy and the final output format. To further improve reasoning and exploration, \citet{START} propose START, which introduces a hint-infer mechanism that manually inserts hint strings during inference. This encourages the LLM to self-reflect and make better use of external tools. They also design Hint-RFT, a method that performs rejection sampling and revises reasoning trajectories to support the supervised fine-tuning of search agents. More recently, Atom-Searcher~\cite{atom-searcher} has been proposed to decompose the holistic ``thinking'' process into several finer-grained ``atom-thinking'' actions. These actions are guided by reasoning reward models, which provide precise feedback on reasoning trajectories, going beyond simple outcome-based rewards.

The main advantage of the single-agent framework lies in its simplicity, as it can be trained end-to-end via RL. This allows researchers to explore the reasoning limits of a single model using carefully designed optimization algorithms. However, a single agent often struggles with highly complex queries that require extensive tool use and long-context reasoning. To address this limitation, multi-agent frameworks have been explored, where multiple agents collaborate to complete complex search and reasoning tasks.



\subsubsection{Multi-Agent Frameworks}
Multi-agent frameworks utilize multiple, specialized LLMs that collaborate to complete complex search and reasoning tasks. This approach enables a division of labor, where each agent focuses on a distinct function, leading to improved system effectiveness and efficiency. For example, KwaiAgents~\cite{KwaiAgents} separates reasoning and summarization into two distinct agents. The reasoning agent is equipped with capabilities such as query understanding, external documents referencing, memory management, and task execution via a hybrid search–browse toolkit. Similarly, \citet{AI-SearchPlanner} propose decoupling the search agent into a search planner and a generator, with optimization efforts primarily on the planner. Building on this, MindSearch~\cite{MindSearch} introduces a multi-agent framework inspired by human cognitive processes for web retrieval. Its architecture consists of a \textit{WebPlanner} and multiple \textit{WebSearchers}. The WebPlanner acts as a high-level controller, decomposing user queries into atomic sub-questions and designing the reasoning process. The WebSearchers then perform hierarchical retrieval over the web, guided by these sub-questions. By employing a coarse-to-fine selection strategy, they efficiently filter valuable information from a large pool of web pages, thereby alleviating the information overload that often hinders LLMs. Alita~\cite{Alita} advances this idea by incorporating self-evolving capabilities through a dynamic MCP box. Its manager agent handles central task planning and MCP brainstorming, deciding whether to generate new MCP tools for emerging tasks. The web agent, in turn, is responsible for browsing and retrieving external information. Similarly, OWL~\cite{OWL} further decouples central planning and task execution to improve generalization across different domains. OWL consists of a domain-agnostic planner agent for high-level task decomposition, a coordinator agent for managing task assignments and dependencies, and a set of specialized agents with domain-specific toolkits that execute subtasks and report results. This modular design allows researchers to focus optimization efforts on the planner, enhancing its adaptability while minimizing training complexity for other components. 

The primary advantage of the multi-agent framework is that it allows individual agents to specialize in distinct tasks, enhancing overall system effectiveness and efficiency. However, it introduces challenges in jointly optimizing multiple agents through RL.  Current approaches often limit optimization to the core planner agent, which plays the most critical role in coordinating the framework. Therefore, developing more stable and efficient RL strategies for multi-agent search systems remains a promising direction for future research.

\subsection{Information Seeking Module}\label{sec:sa_tool}
To handle atomic queries or complete sub-tasks from an upstream planner, a search agent typically relies on an information-seeking module that locates, collects, and synthesizes relevant information. These approaches can be roughly categorized into two types: API-based and browsing-based methods. API-based methods use search engine APIs to retrieve information, while browsing-based methods construct sandboxes or virtual environments that enable agents to simulate human-like web interactions. 

\subsubsection{API-based Information Seeking}
The most straightforward information-seeking strategy is to leverage search engine and scientific database APIs as external tools. Many commercial applications, such as Gemini DeepResearch and Grok DeepSearch,\footnote{Gemini DeepResearch: \url{https://gemini.google/overview/deep-research/}, Grok DeepSearch: \url{https://x.ai/news/grok-3}}
rely on APIs like Google Search, Bing Search, and X Search to access external knowledge. In the research community, Cognitive Kernel-Pro~\cite{Cognitive-Kernel-Pro} uses the free DuckDuckGo search interface to create a fully open-source pipeline, while CoSearch-Agent~\cite{co-search-agent} integrates SerpApi for real-time search within Slack-based environments. Beyond basic search APIs, other systems incorporate specialized APIs to refine the retrieval process. For example, Search-o1~\cite{search-o1} and Agent Laboratory~\cite{Agent-Laboratory} use Jina Reader API to extract and refine web passages for downstream reasoning, and the arXiv API to obtain academic metadata. Similarly, AI Scientist~\cite{AI-scientist} employs the Semantic Scholar API to verify citation relationships. While simple and accessible, API-based approaches often struggle with complex, dynamic content rendered by JavaScript, interactive components, or information gated behind authentication. 

\subsubsection{Browsing-based Information Seeking}
 In contrast to API-based methods, browsing-based information-seeking approaches provide search agents with interactive environments that simulate human-web interactions. For example, Manus AI's browsing agent creates a sandboxed Chromium instance for each subtask,\footnote{Manus AI: \url{https://manus.im/}} while AutoGLM~\footnote{AutoGLM: \url{https://autoglm-research.zhipuai.cn/}} sequentially opens web pages, reads content, and generates refined reports. In research, AutoAgent~\cite{autoagent} uses the BrowserGym environment to perform scrolling and interaction with webpage components. SimpleDeepSearcher~\cite{simpledeepsearcher} and Tool-star~\cite{toolstar} further compress retrieved content from both browsing and API-based extraction to generate condensed references for answer generation. While browsing-based approaches are better suited for retrieving real-time and deeply nested content, they typically incur higher latency and resource costs. To address the high costs associated with real search APIs, some recent works explore alternative approaches. For example, ZeroSearch~\cite{zerosearch} trains LLMs to simulate search engine behavior without making actual API calls, thereby significantly reducing training costs. Additionally, some methods, such as Alita~\cite{Alita}, propose to dynamically create new MCP tools during the agent's reasoning process, enabling a self-evolving capability that reduces reliance on pre-defined toolkits and further optimizes computation costs.

\subsection{Optimization}\label{sec:sa_tuning}
To transform general-purpose LLMs into specialized search agents, researchers have explored various optimization and fine-tuning methods. These approaches aim to internalize advanced search skills, such as planning, reasoning, and tool usage into the model's parametric knowledge. The ultimate goal is to enable agents to perform exploratory information acquisition. Based on the progressive levels of search capabilities, this section categorizes mainstream agent tuning methods into three paradigms: strategic retrieval optimization, iterative search tuning, and autonomous open-web search.

\subsubsection{Strategic Retrieval Optimization}
In basic search scenarios, agents must first learn when to retrieve information and how to formulate high-quality queries. Early RAG methods typically follow a fixed and passive pipeline: given a question, the system directly performs a search and then generates an answer based on the results. This approach is often inefficient, as it can lead to redundant searches and struggles to handle irrelevant information. 

Recent studies have introduced strategic retrieval optimization techniques that enable agents to explicitly model retrieval decisions, thereby balancing search costs against the potential benefit of new information. For example, Open-RAG~\cite{Open-RAG} proposes a ``hybrid adaptive retrieval'' mechanism that learns to generate specific tokens to control its retrieval behavior. This work also introduces a constructive learning paradigm, which actively injects distracting information into training data to enhance the model's robustness and discrimination capabilities against noisy or irrelevant search results. Similarly, DeepRAG~\cite{DeepRAG} models retrieval decisions as a Markov decision process, using imitation learning to train models to weigh the benefits of relying on internal knowledge versus performing an external search at each reasoning step. This provide the model with the dynamic ability to decide when to retrieve. ATLAS~\cite{ATLAS-Agent} takes a different approach by applying gradient backpropagation only on ``critical steps'' within expert trajectories. In search tasks, these steps correspond to key decisions such as initiating a search or formulating a core query. By focusing the training signal on these strategic points, this method improves the agent's core decision-making and generalization ability.

\subsubsection{Iterative Retrieval Tuning}
For complex tasks that cannot be solved with a single retrieval, search agents require multi-step reasoning and iterative information acquisition. These capabilities are crucial for applications such as multi-hop question answering and open-domain problem solving, where agents must converge on an answer through a dynamic cycle of ``think-search-integrate-rethink''. Research in this area has leveraged both supervised and reinforcement learning paradigms.

Supervised learning trains models by providing expert trajectories that include all intermediate reasoning and retrieval steps. CoRAG~\cite{CoRAG} exemplifies this approach by automatically generating retrieval chains with intermediate sub-queries and sub-answers for existing datasets. Through rejection sampling, it enables models to explicitly learn multi-step retrieval patterns. Similarly, Auto-RAG~\cite{Auto-RAG} focuses on synthesizing instruction data that contains retrieval decision processes, allowing models to master autonomous multi-round retrieval logic through SFT.

Reinforcement learning allows agents to autonomously explore and learn optimal dynamic search strategies through interaction with an environment. Works such as ReSearch~\cite{ReSearch}, R1-Searcher~\cite{R1-Searcher}, and Search-R1~\cite{Search-R1} all adopt RL frameworks, defining search as a learnable action within the reasoning process. Agents learn when and how to intersperse search queries by maximizing rewards from task success. Among these, R1-Searcher designs a two-stage RL training pipeline that effectively decouples the objectives of learning to use tools from learning to solve problems with tools. At the algorithmic level, ARPO~\cite{ARPO} optimizes training efficiency for iterative agents by proposing an entropy-based adaptive exploration mechanism. This method increases exploration intensity at critical decision points where models show high uncertainty, significantly reducing training costs.

\subsubsection{Autonomous Open-Web Search}
At the highest level, search agents must be able to operate autonomously within open and dynamic web environments. This capability requires models to handle complex challenges such as noisy data, conflicting information from multiple sources, and a lack of explicit supervision. To succeed, they must possess advanced skills in information planning, cross-validation, and multimodal understanding.

Recent studies show that end-to-end RL is an effective path for endowing models with autonomous research capabilities. WebAgent-R1~\cite{WebAgent-R1} and DeepResearcher~\cite{DeepResearcher} use sparse reward training in real browser environments, enabling agents to autonomously plan search paths, verify information, and integrate knowledge from various sources. WebThinker~\cite{WebThinker} proposes an ``Think-Search-and-Draft'' strategy, using iterative online direct preference optimization (DPO) to enable agents to seamlessly switch between information collection, reasoning, and content generation.

Subsequent research has focused on building more systematic training methodologies. Works such as WebDancer~\cite{WebDancer}, WebSailor~\cite{WebSailor}, and WebShaper~\cite{WebShaper} demonstrate that a combination of high-quality data synthesis and hybrid training strategies is an effective path for training advanced agents. These works have made important innovations at the data level. For example, WebShaper proposes a ``formalization-driven'' data synthesis framework that generates logically consistent data from a task's reasoning structure. WebWatcher~\cite{WebWatcher} further advances this filed by incorporating visual information during training, enabling models to understand and utilize both images and texts on web pages, thereby moving toward human-like research capabilities. Through these methods, search agents are gradually evolving into research-oriented agents capable of autonomous exploration and information integration on the open web.

\subsection{Benchmarks and Resources}\label{sec:sa_data}
To effectively evaluate and advance search agents, a diverse set of benchmarks and resources is essential. Current evaluation methodologies can be broadly categorized into two main paradigms: QA-style benchmarks, which assess the agent's ability to answer complex questions, and task-oriented benchmarks, which measure its capacity for planning, tool use, and environmental interaction. Additionally, a growing number of agent platforms and datasets serve as valuable resources for both evaluation and model training.

\subsubsection{QA Benchmarks}
QA benchmarks are designed to evaluate problem-solving and reasoning capabilities of search agents. They range from simple factual recall to multi-hop reasoning and expert-level challenges. 

\mypara{Single-hop QA.} It involves questions that can be answered by retrieving information from a single document or source. These benchmarks, such as TriviaQA~\cite{TriviaQA}, SimpleQA~\cite{SimpleQA}, PopQA~\cite{PopQA}, and Natural Questions (NQ)~\cite{nq}, primarily test a model's ability to perform open-domain factual retrieval and reading comprehension. For example, NQ provides anonymized Google search queries paired with human-annotated answers and Wikipedia evidence.

\mypara{Multi-hop QA.} It requires models to reason over and combine information from multiple sources to find the answer. A typical multi-hop QA dataset is HotpotQA~\cite{HotpotQA}. It focuses on multi-hop reasoning by providing questions that require chaining evidence across multiple Wikipedia pages. 2WikiMultiHopQA~\cite{2WikiMultiHopQA} extends this by mixing structured knowledge and unstructured text and providing explicit reasoning paths for a more fine-grained evaluation of multi-step inference.

\mypara{Expert-level Challenges.} They are designed to be extremely difficult, often requiring deep domain knowledge and complex reasoning to push the limits of advanced models. Humanity's Last Exam (HLE)~\cite{HLE} assembles thousands of hard, expert-crafted questions to stress test models across broad domains. Benchmarks like BrowseComp~\cite{BrowseComp} attempt to force genuine web-based retrieval by filtering out items solvable from parametric memory. However, even with these efforts, top-performing systems can still exploit internal knowledge, which may overstate their true research capability.

\subsubsection{Task-oriented Benchmarks}
Task-oriented benchmarks assess an agent's practical skills in planning, tool use, and interaction with various environments.

\mypara{General Assistant Workflows.} GAIA~\cite{GAIA}, AssistantBench~\cite{AssistantBench}, and Magnetic-One~\cite{Magentic_One} cover broad assistant tasks that require planning across dialogue, retrieval, and simple tool calls. GAIA, for instance, measures an agent's end-to-end task management, while Magnetic-One emphasizes robustness across diverse domains and chained subtasks.

\mypara{Code and Research.} SWE-bench~\cite{SWE-bench}, HumanEvalFix~\cite{HumanEvalFix}, MLE-bench~\cite{MLE-bench}, and MLAgentBench~\cite{MLAgentBench} probe pipelines centered on software engineering and research. They require agents to perform tasks like code implementation, debugging, experiment setup, and hyperparameter tuning.

\mypara{Multi-Agent Coordination.} RE-Bench~\cite{RE-Bench} and RESEARCHTOWN~\cite{RESEARCHTOWN} stress multi-agent collaboration, role assignment, and iterative refinement on shared research goals.

\mypara{GUI Control.} WebArena~\cite{agent/dynamic/webarena} and SpaBench~\cite{SpaBench} extend evaluation to include direct interface manipulation, measuring an agent's ability to control web UIs or simulated devices and handle noisy, stateful environments.



\subsubsection{Agent Datasets and Platforms}
Beyond static benchmarks, several projects offer comprehensive resources that bundle evaluation suites with data-generation pipelines and agent toolkits. These resources serve as both benchmarks and valuable training corpora for agent research.

The Alibaba-NLP WebAgent repository is a notable example, packaging a web-traversal benchmark (WebWalkerQA~\cite{WebWalker}) with agent models and data tools (WebDancer~\cite{WebDancer}, WebShaper~\cite{WebShaper}, and WebSailor~\cite{WebSailor}). Specifically, WebWalkerQA~\cite{WebWalker} probes an agent's ability to traverse sites and extract evidence across multiple subpages, emphasizing structured navigation over single-turn retrieval. WebDancer~\cite{WebDancer} implements a four-stage training paradigm and releases both models and browsing trajectories, enabling reproducible, end-to-end evaluation. WebShaper~\cite{WebShaper} provides a ``formalization-driven'' data synthesis pipeline that systematically generates information-seeking instances, making it valuable for cold-starting agents and for studying data-centric training strategies. The recent model releases from WebSailor~\cite{WebSailor} demonstrate how post-training and specialized agent tuning can yield stronger navigation and planning behaviors on these benchmarks.

%% file: S8-future-direction.tex
\section{Future Direction}
\label{sec:future}
In this survey, we comprehensively reviewed recent advancements in LLM-enhanced IR systems and discussed their limitations. Since the integration of LLMs into IR systems is still in its early stages, there are still many opportunities and challenges. In this section, we summarize the potential future directions in terms of the four modules in an IR system we just discussed, namely query rewriter, retriever, reranker, and reader. In addition, as evaluation has also emerged as an important aspect, we will also introduce the corresponding research problems that need to be addressed in the future. Another discussion about important research topics on applying LLMs to IR can be found in a recent perspective paper~\cite{ir_perspective}. 

\subsection{Query Rewriter}
LLMs have enhanced query rewriter for both ad-hoc and conversational search scenarios. Most of the existing methods rely on prompting LLMs to generate new queries. While yielding remarkable results, the refinement of rewriting quality and the exploration of potential application scenarios require further investigation.

$\bullet$ \textit{Rewriting queries according to ranking performance.}
A typical paradigm of prompting-based methods is providing LLMs with several ground-truth rewriting cases (optional) and the task description of query rewriter. Despite LLMs being capable of identifying potential user intents of the query~\cite{intent5}, they lack awareness of the resulting retrieval quality of the rewritten query. The absence of this connection can result in rewritten queries that seem correct yet produce unsatisfactory ranking results. Although some existing studies have used reinforcement learning to adjust the query rewriter process according to generation results~\cite{ma2023query}, a substantial realm of research remains unexplored concerning the integration of ranking results. 

$\bullet$ \textit{Improving query rewriter in conversational search.}
As yet, primary efforts have been made to improve query rewriter in ad-hoc search. In contrast, conversational search presents a more developed landscape with a broader scope for LLMs to contribute to query understanding. By incorporating historical interactive information, LLMs can adapt system responses based on user preferences, providing a more effective conversational experience. However, this potential has not been explored in depth. In addition, LLMs could also be used to simulate user behavior in conversational search scenarios, providing more training data, which are urgently needed in current research.

$\bullet$ \textit{Achieving personalized query rewriter.}
LLMs offer valuable contributions to personalized search through their capacity to analyze user-specific data. In terms of query rewriter, with the excellent language comprehension ability of LLMs, it is possible to leverage them to build user profiles based on users' search histories (\eg, issued queries, click-through behaviors, and dwell time). This empowers the achievement of personalized query rewriter for enhanced IR and finally benefits personalized search or personalized recommendation.

\subsection{Retriever}
Leveraging LLMs to improve retrieval models has received considerable attention, promising an enhanced understanding of queries and documents for improved ranking performance. However, despite strides in this field, several challenges and limitations still need to be investigated in the future:

$\bullet$ \textit{Reducing the latency of LLM-based retrievers.} LLMs, with their massive parameters and world knowledge, often entail high latency during the inferring process. This delay poses a significant challenge for practical applications of LLM-based retrievers, as search engines require in-time responses. To address this issue, promising research directions include transferring the capabilities of LLMs to smaller models, exploring quantization techniques for LLMs in IR tasks, and so on.

$\bullet$ \textit{Simulating realistic queries for data augmentation.} Since the high latency of LLMs usually blocks their online application for retrieval tasks, many existing studies have leveraged LLMs to augment training data, which is insensitive to inference latency. Existing methods that leverage LLMs for data augmentation often generate queries without aligning them with real user queries, leading to noise in the training data and limiting the effectiveness of retrievers. As a consequence, exploring techniques such as reinforcement learning to enable LLMs to simulate the way that real queries are issued holds the potential for improving retrieval tasks.

$\bullet$ \textit{Incremental indexing for generative retrieval.} As elaborated in Section~\ref{gen_ir}, the emergence of LLMs has paved the way for generative retrievers to generate document identifiers for retrieval tasks. This approach encodes document indexes and knowledge into the LLM parameters. However, the static nature of LLM parameters, coupled with the expensive fine-tuning costs, poses challenges for updating document indexes in generative retrievers when new documents are added. Therefore, it is crucial to explore methods for constructing an incremental index that allows for efficient updates in LLM-based generative retrievers. 

$\bullet$ \textit{Supporting multi-modal search.} Web pages usually contain multi-modal information, including texts, images, audios, and videos. However, existing LLM-enhanced IR systems mainly support retrieval for text-based content. A straightforward solution is to replace the backbone with multi-modal large models, such as GPT-4~\cite{gpt-4}. However, this undoubtedly increases the cost of deployment. A promising yet challenging direction is to combine the language understanding capability of LLMs with existing multi-modal retrieval models. By this means, LLMs can contribute their language skills in handling different types of content.

\subsection{Reranker}
In Section~\ref{sec:rank}, we have discussed the recent advanced techniques of utilizing LLMs for the reranking task. Some potential future directions in reranking are discussed as follows.

$\bullet$ \textit{Enhancing the online availability of LLMs.} Though effective, many LLMs have a massive number of parameters, making it challenging to deploy them in online applications. Besides, many reranking methods~\cite{sun2023chatgpt, ma2023zero} rely on calling LLM APIs, incurring considerable costs. Consequently, devising effective approaches (such as distilling to small models) to enhance the online applicability of LLMs emerges as a research direction worth exploring.

$\bullet$ \textit{Improving personalized search.} Many existing LLM-based reranking methods mainly focus on the ad-hoc reranking task. However, by incorporating user-specific information, LLMs can also improve the effectiveness of the personalized reranking task. For example, by analyzing users' search history, LLMs can construct accurate user profiles and rerank the search results accordingly, providing personalized results with higher user satisfaction.

$\bullet$ \textit{Adapting to diverse ranking tasks.} In addition to document reranking, there are also other ranking tasks, such as response ranking, evidence ranking, entity ranking and etc., which also belong to the universal information access system. Navigating LLMs towards adeptness in these diverse ranking tasks can be achieved through specialized methodologies, such as instruction tuning. Exploring this avenue holds promise as an intriguing and valuable research trajectory.

\subsection{Reader}
With the increasing capabilities of LLMs, the future interaction between users and IR systems will be significantly changed. Due to the powerful natural language processing and understanding capabilities of LLMs, the traditional search paradigm of providing ranking results is expected to be progressively replaced by synthesizing conclusive answering passages for user queries using the reader module. Although such strategies have already been investigated by academia and facilitated by industry as we stated in Section~\ref{sec:reader}, there still exists much room for exploration.

$\bullet$ \textit{Improving the reference quality for LLMs.} To support answer generation, existing approaches usually directly feed the retrieved documents to the LLMs as references. However, since a document usually covers many topics, some passages in it may be irrelevant to the user queries and can introduce noise during LLMs' generation. Therefore, it is necessary to explore techniques for extracting relevant snippets from retrieved documents, enhancing the performance of retrieval-augmented generation.

$\bullet$ \textit{Improving the answer reliability of LLMs.} Incorporating the retrieved references has significantly alleviated the ``hallucination'' problem of LLMs. However, it remains uncertain whether the LLMs refer to these supported materials during answering queries. Some studies~\cite{LLMbond} have revealed that LLMs can still provide unfaithful answers even with additional references. Therefore, the reliability of the conclusive answers might be lower compared to the ranking results provided by traditional IR systems. It is essential to investigate the influence of these references on the generation process, thereby improving the credibility of reader-based novel IR systems.

\subsection{Search Agent}
With the outstanding performance of LLMs, the patterns of searching may completely change from traditional IR systems to autonomous search agents.
In Section~\ref{sec:agent}, we have discussed many existing works that utilize a static or dynamic pipeline to autonomously browse the web.
These works are believed to be the pioneering works of the new searching paradigm.
However, there is still plenty of room for further improvements.

$\bullet$ \textit{Enhancing the Trustworthiness of LLMs.}
When LLMs are enabled to browse the web, it is important to ensure the validity of retrieved documents. Otherwise, the unfaithful information may increase the LLMs' hallucination problem. 
Besides, even if the gathered information has high quality, it remains unclear whether they are really used for synthesizing responses.
A potential strategy to address this issue is enabling LLMs to autonomously validate the documents they scrape. 
This self-validation process could incorporate mechanisms for assessing the credibility and accuracy of the information within these documents.

$\bullet$ \textit{Mitigating Bias and Offensive Content in LLMs.} 
The presence of biases and offensive content within LLM outputs is a pressing concern. 
This issue primarily stems from biases inherent in the training data and will be amplified by the low-quality information gathered from the web.
Achieving this requires a multi-faceted approach, including improvements in training data, algorithmic adjustments, and continuous monitoring for bias and inappropriate content that LLMs collect and generate.

\subsection{Evaluation}
LLMs have attracted significant attention in the field of IR due to their strong ability in context understanding and text generation. To validate the effectiveness of LLM-enhanced IR approaches, it is crucial to develop appropriate evaluation metrics. Given the growing significance of readers as integral components of IR systems, the evaluation should consider two aspects: assessing ranking performance and evaluating generation performance.

$\bullet$ \textit{Generation-oriented ranking evaluation.} Traditional evaluation metrics for ranking primarily focus on comparing the retrieval results of IR models with ground-truth (relevance) labels. Typical metrics include precision, recall, mean reciprocal rank (MRR)~\cite{MRR}, mean average precision (MAP), and normalized discounted cumulative gain (nDCG)~\cite{NDCG}. These metrics measure the alignment between ranking results and human preference on using these results. Nevertheless, these metrics may fall short in capturing a document's role in the generation of passages or answers, as their relevance to the query alone might not adequately reflect this aspect. This effect could be leveraged as a means to evaluate the usefulness of documents more comprehensively. A formal and rigorous evaluation metric for ranking that centers on generation quality has yet to be defined.

$\bullet$ \textit{Text generation evaluation.} The wide application of LLMs in IR has led to a notable enhancement in their generation capability. Consequently, there is an imperative demand for novel evaluation strategies to effectively evaluate the performance of passage or answer generation. Previous evaluation metrics for text generation have several limitations, including: (1)~Dependency on lexical matching: methods such as BLEU~\cite{bleu} or ROUGE~\cite{rouge} primarily evaluate the quality of generated outputs based on $n$-gram matching. This approach cannot account for lexical diversity and contextual semantics. As a result, models may favor generating common phrases or sentence structures rather than producing creative and novel content. (2)~Insensitivity to subtle differences: existing evaluation methods may be insensitive to subtle differences in generated outputs. For example, if a generated output has minor semantic differences from the reference answer but is otherwise similar, traditional methods might overlook these nuanced distinctions. (3)~Lack of ability to evaluate factuality: LLMs are prone to generating ``hallucination'' problems~\cite{DBLP:journals/corr/abs-2303-08896,webbrain,DBLP:journals/corr/abs-2305-11747,DBLP:conf/emnlp/ChenDBQWCW23}. The hallucinated texts can closely resemble the oracle texts in terms of vocabulary usage, sentence structures, and patterns, while having non-factual content. Existing methods are hard to identify such problems, while the incorporation of additional knowledge sources such as knowledge bases or reference texts could potentially aid in addressing this challenge.

\subsection{Bias}
Since ChatGPT was released, LLMs have drawn much attention from both academia and industry. The wide applications of LLMs have led to a notable increase in content on the Internet that is not authored by humans but rather generated by these language models. However, as LLMs may hallucinate and generate non-factual texts, the increasing number of LLM-generated contents also brings worries that these contents may provide fictitious information for users across IR systems. More severely, researchers~\cite{sourcebias1,sourcebias2} show that some modules in IR systems such as retriever and reranker, especially those based on neural models, may prefer LLM-generated documents, since their topics are more consistent and the perplexity of them are lower compared with human-written documents. The authors refer to this phenomenon as the ``source bias'' towards LLM-generated text. It is challenging but necessary to consider how to build IR systems free from this category of bias.

%% file: S9-conclusion.tex
\section{Conclusion}
\label{sec:conclu}
In this survey, we have conducted a thorough exploration of the transformative impact of LLMs on IR across various dimensions. We have organized existing approaches into distinct categories based on their functions: query rewriter, retrieval, reranking, and reader modules. In the domain of query rewriter, LLMs have demonstrated their effectiveness in understanding ambiguous or multi-faceted queries, enhancing the accuracy of intent identification. In the context of retrieval, LLMs have improved retrieval accuracy by enabling more nuanced matching between queries and documents, considering context as well. Within the reranking realm, LLM-enhanced models consider more fine-grained linguistic nuances when re-ordering results. The incorporation of reader modules in IR systems represents a significant step towards generating comprehensive responses instead of mere document lists. The integration of LLMs into IR systems has brought about a fundamental change in how users engage with information and knowledge. From query rewriter to retrieval, reranking, and reader modules, LLMs have enriched each aspect of the IR process with advanced linguistic comprehension, semantic representation, and context-sensitive handling. As this field continues to progress, the journey of LLMs in IR portends a future characterized by more personalized, precise, and user-centric search encounters.

This survey focuses on reviewing recent studies of applying LLMs to different IR components and using LLMs as search agents. Beyond this, a more significant problem brought by the appearance of LLMs is: is the conventional IR framework necessary in the era of LLMs? For example, traditional IR aims to return a ranking list of documents that are relevant to issued queries. However, the development of generative language models has introduced a novel paradigm: the direct generation of answers to input questions. Furthermore, according to a recent perspective paper~\cite{ir_perspective}, IR might evolve into a fundamental service for diverse systems. For example, in a multi-agent simulation system~\cite{DBLP:journals/corr/abs-2304-03442}, an IR component can be used for memory recall. This implies that there will be many new challenges in future IR. 